\documentclass[10pt,twocolumn,letterpaper]{article}

\usepackage{cvpr}              
\usepackage{amsmath}
\usepackage{booktabs}       
\usepackage{nicefrac}       
\usepackage{microtype}      
\usepackage{multirow}
\usepackage{tabularx}
\usepackage{adjustbox}
\usepackage{times}  
\usepackage{helvet}  
\usepackage{courier}  
\usepackage[hyphens]{url}  
\usepackage{graphicx} 
\usepackage{caption} 
\usepackage{algorithm}
\usepackage{algorithmic}
\usepackage{newfloat}
\usepackage{listings}
\usepackage{xcolor} 
\definecolor{myorange}{RGB}{255,128,0}
\definecolor{myblue}{RGB}{68,114,196}
\definecolor{myred}{RGB}{255,0,0}

\definecolor{cvprblue}{rgb}{0.21,0.49,0.74}
\usepackage[pagebackref,breaklinks,colorlinks,citecolor=cvprblue]{hyperref}

\def\paperID{15244} 
\def\confName{CVPR}
\def\confYear{2024}

\title{RetouchUAA: Unconstrained Adversarial Attack via Image Retouching}

\author{Mengda Xie\\
Guangzhou University, China\\
{\tt\small 1112206011@e.gzhu.edu.cn}
\and
Yiling He\\
Zhejiang University, China\\
{\tt\small yilinghe@zju.edu.cn}
\and
Meie Fang~\thanks{Corresponding author}\\
Guangzhou University, China\\
{\tt\small fme@gzhu.edu.cn}
}

\begin{document}
\maketitle

\begin{abstract}
    Deep Neural Networks (DNNs) are susceptible to adversarial examples. 
    Conventional attacks generate controlled noise-like perturbations that fail to reflect real-world scenarios and hard to interpretable. In contrast, recent unconstrained attacks mimic natural image transformations occurring in the real world for perceptible but inconspicuous attacks, yet compromise realism due to neglect of image post-processing and uncontrolled attack direction.
    In this paper, we propose RetouchUAA, an unconstrained attack that exploits a real-life perturbation: image retouching styles, highlighting its potential threat to DNNs.
    Compared to existing attacks, RetouchUAA offers several notable advantages.   
    Firstly, RetouchUAA excels in generating interpretable and realistic perturbations through two key designs: the image retouching attack framework and the retouching style guidance module.  
    The former custom-designed human-interpretability retouching framework for adversarial attack by linearizing images while modelling the local processing and retouching decision-making in human retouching behaviour, provides an explicit and reasonable pipeline for understanding the robustness of DNNs against retouching.
    The latter guides the adversarial image towards standard retouching styles, thereby ensuring its realism.
    Secondly, attributed to the design of the retouching decision regularization and the persistent attack strategy, RetouchUAA also exhibits outstanding attack capability and defense robustness, posing a heavy threat to DNNs.
    Experiments on ImageNet and Place365 reveal that RetouchUAA achieves nearly 100\% white-box attack success against three DNNs, while achieving a better trade-off between image naturalness, transferability and defense robustness than baseline attacks.
\end{abstract}

\section{Introduction}\label{sec1}
\par Deep Neural Networks (DNNs) excel in computer vision tasks like image recognition~\cite{he2016deep,liu2022convnet}, object tracking~\cite{song2022transformer}, and image generation~\cite{karras2018progressive}. However, they are inherently vulnerable to adversarial attacks, where inconspicuous input perturbations can cause incorrect model outputs~\cite{szegedy2013intriguing}.
\begin{figure}
 \vspace{-0.0cm}
  \includegraphics[width=1\linewidth]{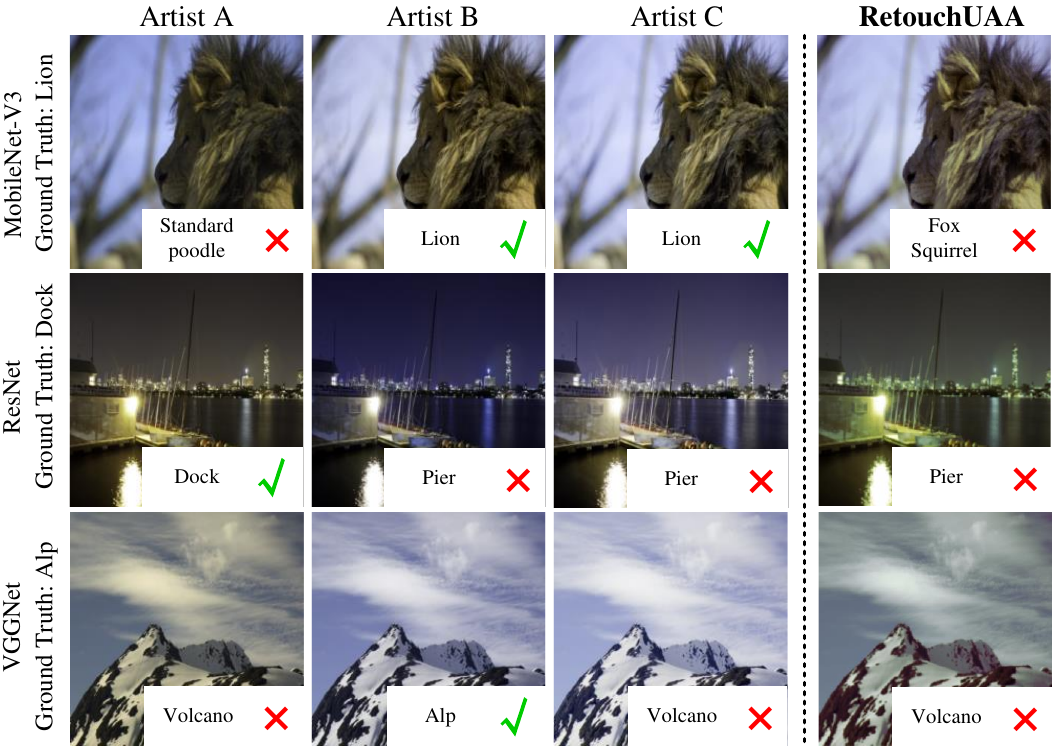}
  \caption{Impact of various retouching styles on pretrained models: the first three columns display artist retouching, the last column adversarial retouching by our method. Bottom right of each image shows classification results, crosses denote misclassifications.}
\vspace{-0.7cm}
  \label{fig1}
\end{figure}
\par Early adversarial methods~\cite{goodfellow2014explaining, kurakin2018adversarial, madry2018towards, carlini2017towards}, seeking noise-type perturbations, are limited to small perturbations to retain similarity to original samples. However, these noise patterns are difficult to interpret and seldom threaten real systems as they rarely occur in reality~\cite{zhao2020adversarial2}.
Recent unconstrained attacks offer greater flexibility and realism by simulating physical transformations like rain, haze~\cite{gao2021advhaze}, blur~\cite{guo2020watch}, and vignetting~\cite{tian2021ava}, revealing the detrimental effects of these physical factors on DNNs. Other unconstrained attacks, such as AdvCF~\cite{zhao2020adversarial1, zhao2020adversarial2}, SAE~\cite{hosseini2018semantic} and ALA~\cite{sun2022ala}, make alterations to natural image attributes like color or brightness, showing promising attack results. These modifications, resembling user retouching more than malicious attacks, tend to
evade suspicion effectively. However, they only consider certain specific operations in the retouching process, hence constraining their attack strength and image naturalness due to stringent attack direction limitations. Crucially, these methods do not truly simulate the variations of retouching style that occur in the real world, as they overlook intricate post-processing steps inherent in human retouching, such as local adjustments and retouching decisions, while lacking guidance from natural images for the generated adversarial images. Consequently, it is unclear the impact of real-world retouching styles on DNNs. Moreover, the potential for the retouching style as a new attack perspective and whether we can effectively defend against it is also uncertain.

\par Considering these factors, in this paper, we explore the impact of image retouching style on DNN models from an adversarial attack perspective. 
We illustrate the effectiveness of the retouching style attack in Figure~\ref{fig1} from two aspects.   
Firstly, retouching styles are widely accepted by humans, making style alterations less likely to arouse suspicion. As shown in the columns one to three of Figure~\ref{fig1}, the images from the MIT-Adobe FiveK dataset~\cite{bychkovsky2011learning}, retouched by different photographers, both look natural to humans. Second, variations in retouching style could significantly impact model performance: the evaluation of three popular computer vision architectures (MobileNet, ResNet, and VGGNet) in Figure~\ref{fig1} reveals the problem that models may produce inaccurate results when the retouching style changes.

\par
\begin{figure*}[t]
    \centering
    \includegraphics[width=0.95\textwidth]{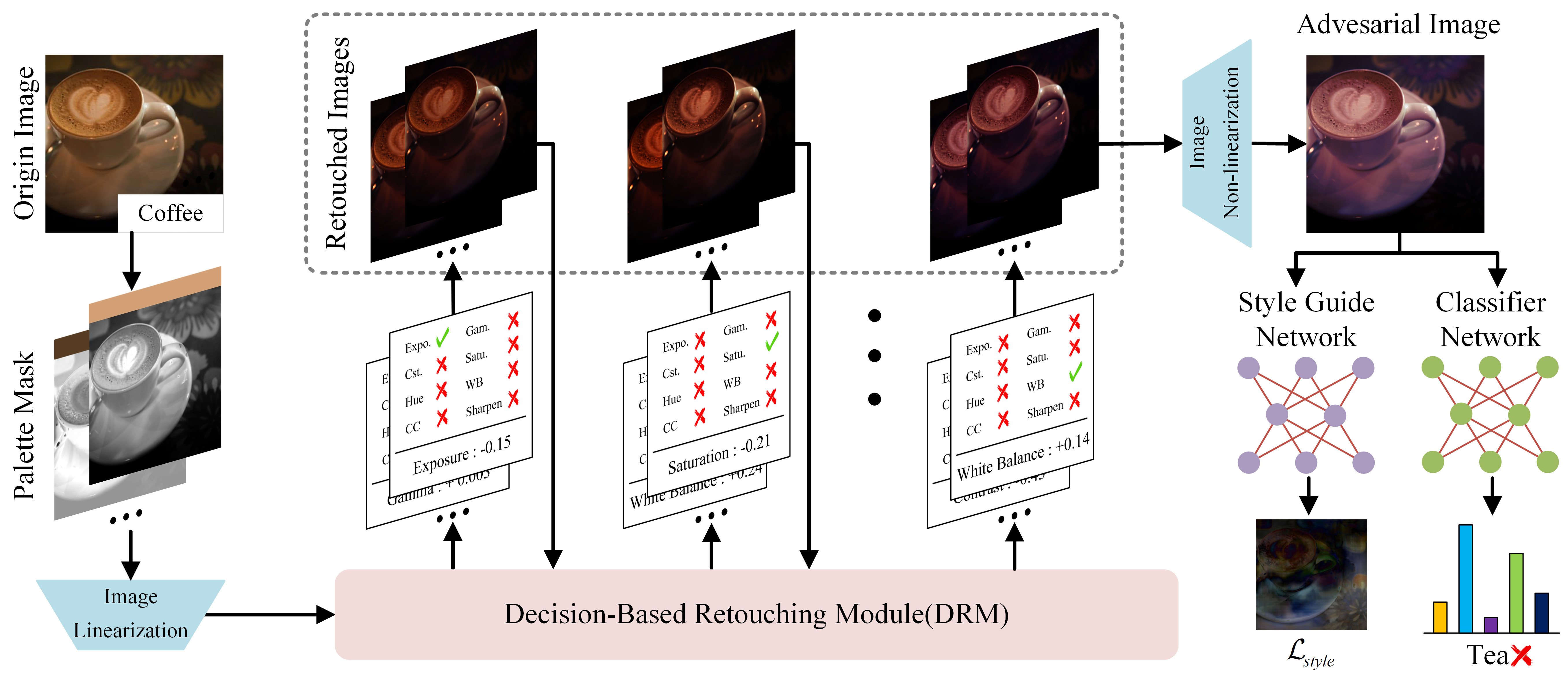}
    \vspace{-0.2cm}
    \caption{RetouchUAA's workflow. The image undergoes soft mask generation and linearization. Then, DRM optimizes retouching sequences and parameters for different mask areas based on style and visual task losses (e.g., classification), producing natural yet aggressive retouched images to mislead visual task models.}
    \label{fig2}
    \vspace{-0.6cm}
\end{figure*}

 Motivated by the above initial observation, we developed RetouchUAA to study retouching styles' effects on DNNs, depicted in Figure~\ref{fig2}. To generate adversarial images that conform to the retouching pipeline, we designed a human-interpretability retouching-based attack framework that mimics image states in the retouching process as well as human retouching behaviours. Specifically, this framework adopts a linear color space suitable for image retouching, and emulates typical photographic adjustments using soft masks for localized retouching.
Furthermore, we devised a Decision-based Retouching Module (DRM) to emulate human decision-making in retouching and identify adversarial retouch combinations for generating retouched images. In particular, we introduced the Gumbel-Softmax technique to overcome DRM's non-differentiable decision-making challenge.
Considering the above attack framework lacks attack direction constraints, which may result in unnatural adversarial images, we introduced a style guidance module to direct the generation towards standard retouching styles. 
Finally, we proposed retouching decision regularization and a persistent attack strategy to boost RetouchUAA's attack strength and defense robustness. The former encourages retouching operations generated by DRM, broadening attack avenues. The latter strategy further distances adversarial images from the original label's decision boundary while maintaining their naturalness after the initial attack succeeds.

\par The final column of Figure~\ref{fig1} presents adversarial images produced by RetouchUAA, featuring a range of diverse, natural, yet aggressive retouching styles. This underscores the effectiveness of RetouchUAA in emulating retouching styles, while also introducing new challenges for DNNs.

\par Our contributions can be encapsulated as follows.
\begin{itemize}
    \item 
    We propose RetouchUAA, the first comprehensive study of the impact of image retouching on DNNs from the perspective of unconstrained adversarial attacks.
    \item 
    We developed a retouching-based attack framework that simulates image states and human retouching behavior, constrained by a style guidance module for realistic retouching. This aids in analyzing DNNs' robustness against real-world retouching styles.
    \item 
    We designed two components to further explore the attack potential of retouching styles: retouching decision regularization, promoting diversified retouching operations, and a persistent attack strategy to distance adversarial samples from decision boundaries. Both notably improve RetouchUAA's attack strength and defense robustness.
    \item Evaluation results based on two types of visual tasks indicate that DNNs are vulnerable to retouching styles. Further, RetouchUAA exhibits superior attack strength and defense robustness compared to baseline attacks by converting retouching styles, posing a new security challenges for DNNs.

\end{itemize}

\section{Related Work}\label{sec2}

\subsection{Constrained Adversarial Attacks}\label{sec2.1}
Constrained Adversarial Attacks~(CAA) maintains visual similarity between adversarial and original images through bounded perturbations, typically measured by the $L_p$ norm. Given a model $\mathcal{F}$ trained on an image $\boldsymbol{x}$ with truth $y$, and $\mathcal{F}$ can correctly classify $\boldsymbol{x}$. The CAA's goal is to create a minimally perturbed adversarial image $\boldsymbol{x}'$ that causes $\mathcal{F}$ to misclassify $\boldsymbol{x}$ as $y'$:
\begin{equation} \label{eq1}
\begin{gathered}
    \underset{\boldsymbol{x}'}{\mathop{\min }}\,{{\left\| \boldsymbol{x}'-\boldsymbol{x} \right\|}_{p}} \\ 
  s.t.\ \mathcal{F}\left( \boldsymbol{x} \right)=y,\ \mathcal{F}\left( \boldsymbol{x}' \right)=y',\ y\ne y',\ \boldsymbol{x}\in {{\left[ 0,1 \right]}}
\end{gathered}
\end{equation}
where ${{\left\| \boldsymbol{x} \right\|}_{p}}$ represents the $L_p$ norm, quantifying perturbation magnitude. To minimize the perturbation $\eta$, Goodfellow et al.~\cite{goodfellow2014explaining} proposed the Fast Gradient Sign Method (FGSM) with one-step attack; Kurakin et al.~\cite{kurakin2018adversarial} improved it with an iterative gradient descent method. Besides using the $L_p$ norm as a metric, other techniques evaluate similarity between original and adversarial images via visual quality metrics such as perceptual color distance~\cite{zhao2020towards}, psychometric perceptual adversarial similarity score~\cite{wang2004image}, or structural similarity~\cite{rozsa2016adversarial}. It is worth noting that although some of the aforementioned attack methods emphasize visual quality over $L_p$ norm, all strictly control perturbation magnitude, thus are considered constrained attack methods.

\subsection{Unconstrained Adversarial Attacks}\label{sec2.2}
Unconstrained Adversarial Attacks (UAA) aim to create large but non-suspicious perturbations through image transformation like geometric transformations, parameter transformations, and color transformations. Geometric transformations can be basic scaling and rotation~\cite{engstrom2017rotation,kanbak2018geometric}, with recent research~\cite{xiao2018spatially} exploring spatial geometric perturbations as well.
Parameter transformation-based attacks involve altering image generation parameters, demonstrated by Qiu et al.~\cite{qiu2020semanticadv} through editing semantically significant attributes in DNNs. Other methods include perturbing physical parameters like material~\cite{liu2017material}, viewpoint~\cite{athalye2018synthesizing}, relighting~\cite{zhang2021adversarial}, or simulating natural phenomena such as rain~\cite{zhai2022adversarial}, haze~\cite{gao2021advhaze}, blur~\cite{guo2020watch}, vignetting~\cite{tian2021ava}, and shadow~\cite{zhong2022shadows} in image formation.
Finally, color transformation-based attacks modify image colors while preserving texture. Studies like Huang et al.~\cite{sun2022ala} and Afifi et al.~\cite{afifi2019else} show that changing brightness or color temperature impacts DNNs. Zhao et al.\cite{zhao2020adversarial1,zhao2020adversarial2} introduced the Adversarial Color Filter (AdvCF) for color curve manipulation. Laidlaw et al.~\cite{laidlaw2019functional} developed a function for color mapping in adversarial images, while Shamsabadi et al.~~\cite{shamsabadi2020colorfool} and Bhattad et al.~\cite{bhattad2019unrestricted} proposed ColorFool and cAdv, respectively, for targeted color modifications in images.

\section{Methodology}\label{sec3}
\subsection{Problem Formulation}\label{sec3.1}
\par Referring to the constrained adversarial attack, RetouchUAA can be initialized mathematically modelled as follows:
\begin{equation} \label{eq2}
\begin{aligned}
  \underset{\boldsymbol{\theta} }{\mathop{\arg \max }}\,\mathcal{J}\left( \mathcal{F}\left( \mathcal{C}\left( \boldsymbol{x}^{real},\mathrm{ }\boldsymbol{\theta}  \right) \right),y \right)+R\
\end{aligned}
\end{equation}
where $\boldsymbol{x}$ and $y$ denote the clean image and its label. $\mathcal{C}$ encapsulates retouching, $\mathcal{F}$ is the DNN model, and $\mathcal{J}$ denotes the loss function for conventional visual tasks like classification. RetouchUAA's primary goal is searching $\boldsymbol{\theta}$ for $\boldsymbol{x}^{fake}=\mathcal{C}(\boldsymbol{x}^{real}, \boldsymbol{\theta})$, maximizing the task loss $\mathcal{J}(\mathcal{F}(\boldsymbol{x}^{fake}), y)$.

\subsection{Retouching-based Attack Framework}\label{sec3.2}
\par In this section, we propose a retouching-based attack framework with two properties: (1) It simulates the real-world retouching process, allowing us to study the impact of realistic retouching on DNNs. (2) It's differentiable, enabling backpropagation to find adversarial retouching parameters.
We achieve the first property by simulating the linearized retouched images state (Section \ref{sec3.2.1}), along with photographers' retouching actions such as local processing (Section \ref{sec3.2.2}) and operational decision-making (Section \ref{sec3.2.3}). The second property, via a differentiable set of retouching operations adjusting image properties like hue and white balance (Supplementary Material), and employing Gumbel-Softmax for non-differentiable decisions (Sec.~\ref{sec3.2.3}).
\subsubsection{Image Linearization}\label{sec3.2.1}
\par Image retouching typically employs raw-RGB images with linearized scene reference~\cite{afifi2021cie} for their resistance to overexposure and color distortion during retouching, as opposed to ImageNet's nonlinear sRGB images that often degrade in quality when processed directly~\cite{karaimer2016software,nguyen2016raw,afifi2019color}. To counter this, we restore the linear relationship between pixel values and scene radiance by removing gamma.
\begin{equation} \label{eq3}
    \begin{gathered}
    \boldsymbol{x}^{L}=\left\{ \begin{matrix}
       {\boldsymbol{x}^{NL}}/{12.92}\; & if\ \boldsymbol{x}^{NL}\le 0.04  \\
       {{\left( {\left( \boldsymbol{x}^{NL}+0.055 \right)}/{1.06}\; \right)}^{2.4}} & otherwise  \\
    \end{matrix} \right\} \\
    \end{gathered}
\end{equation}
where $NL$ and $L$ denote nonlinear and linearized image states respectively. We observed no added naturalness by removing extra nonlinear operations like tone mapping. After retouching, we apply inverse of Eq.~\ref{eq3} to revert $L$ to $NL$ for visual recognition.

\subsubsection{Palette-Based Soft Mask Generation}\label{sec3.2.2}
\par Traditional retouching strategies typically involve global image operations~\cite{hu2018exposure, yan2014learning}, in contrast to professional photographers who prefer a divide-and-conquer approach for more detailed effects~\cite{moran2020deeplpf}. Adapting to this, we apply Chang et al.'s palette method~\cite{chang2015palette} to detect the main image color and generate $K$ soft masks, $\boldsymbol{\Omega}_{k}$, guiding region-specific retouching in image $I$.
\begin{equation} \label{eq4}
\begin{aligned}
  {{\boldsymbol{x}}_{k+1}}={{\mathcal{R}}_{k}}\left( {{\boldsymbol{x}}_{k}}, {{\boldsymbol{\theta}}_{k}} \right)\odot \boldsymbol{\Omega} {}_{k}+{{\boldsymbol{x}}_{k}}\odot \left( 1-\boldsymbol{\Omega} {}_{k} \right)
\end{aligned}
\end{equation}
where $0\le k<K-1$, $\boldsymbol{x}_{k}$ denotes the retouched image via the $k$-th mask, with $\mathcal{R}_{k}$ and ${\boldsymbol{\theta}}_{k}$ representing the retouching operation and the parameters with the $k$-th mask, respectively. $\boldsymbol{x}_{0}$ represents the initial input image $\boldsymbol{x}^{real}$. The Hadamard product is denoted by $\odot$. $\boldsymbol{\Omega}_{k}$ represents the $k$-th mask generated from the palette, calculated through the Euclidean distance between the image pixels $\boldsymbol{x}$ and the palette $\boldsymbol{C}_k$ within the Lab color space.
\begin{equation} \label{eq5}
\begin{aligned}
{{\boldsymbol{\Omega} }_{k}}\left( x \right)=1-\overline{{{\left\| {{\boldsymbol{x}}^{Lab}}-\boldsymbol{C}_{k}^{Lab} \right\|}_{2}}}
\end{aligned}
\end{equation}
where ${{\left\| \cdot  \right\|}_{2}}$ denotes the $L_2$ norm and $\overline{\cdot}$ means normalization.

\subsubsection{Decision-Based Retouching module}\label{sec3.2.3}
\begin{figure}[b]
  \vspace{-0.4cm}
  \includegraphics[width=1\linewidth]{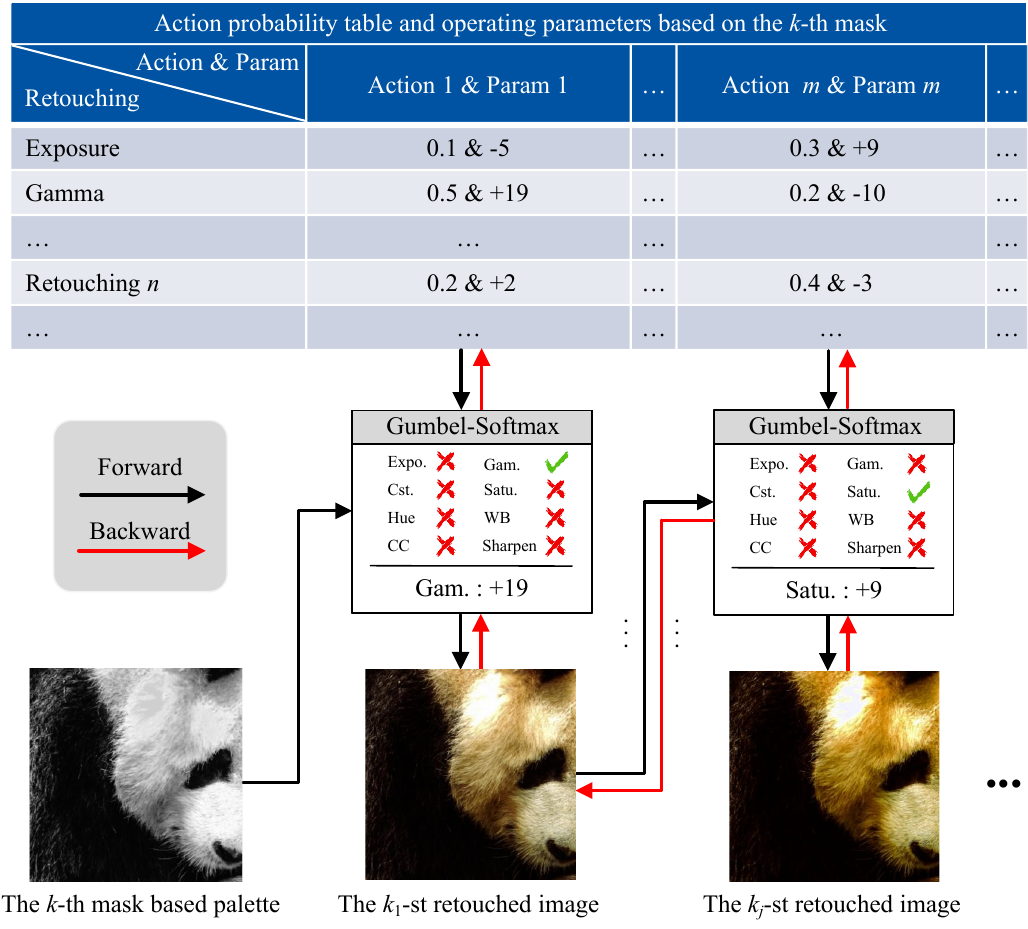}
  \caption{The DRM flowchart. For the $k$-th mask region, at each action, we use Gumbel-Softmax to sample specific retouching operations from the action probability table and obtain the corresponding retouching parameters to retouch the image. The action probability and retouching parameters are updated by backpropagation.}
  \label{fig3}
  \vspace{-0.4cm}
\end{figure}
\par In human retouching, each operation is typically chosen as a discrete decision action, with selected operations performed sequentially. We propose a Decision-Based Retouching Module (DRM) to simulate this behaviour.
Specifically, at each iteration, DRM samples a particular combination of retouching operations from an action probability table and applies them to the image. The values in the action probability table and the corresponding retouching parameters are updated via backpropagation.
We set the total number of actions involved in retouching the $k$-th mask region as $M$. Each action necessitates a choice among $N$ types of retouching operations. $P_{m,n}^{k}$ represents the probability of sampling the $n$-th operation type at the $m$-th action when retouching the $k$-th mask region, according to the action probability table. Here, the corresponding retouching parameters are denoted as ${Z_{m,n}^{k}}$, where $0\le m<M$ and $0\le n<N$.
\par The workflow of DRM is illustrated in Figure~\ref{fig3}. During the retouching process in the $k$-th mask region, we select the $m$-th column of the action probability table as $\boldsymbol{P}_{m}^{k}$ and sample the retouching operation type according to it, thereby obtaining the precise retouching action for the $m$-th action.
\begin{equation} \label{eq6}
\begin{aligned}
  {{\boldsymbol{D}}_{m}^{k}}\leftarrow {{\boldsymbol{P}}_{m}^{k}}
\end{aligned}
\end{equation}
where ${\boldsymbol{D}}_{m}^{k}$ represents the binary decision mask during the $m$-th action for the $k$-th mask region. $D_{m,n}^{k}=1$ means using the $m$-th retouching operation type, while $D_{m,n}^{k}=0$ indicates otherwise. Further, based on $\boldsymbol{D}_{m}^{k}$, we select the corresponding retouching parameter ${{\hat{Z}}_{m}^{k}}$ from the set of retouching parameters ${\boldsymbol{Z}}_{m}^{k}$ for the current retouching operation type:
\begin{equation} \label{eq7}
\begin{aligned}
  {{\hat{Z}}_{m}^{k}}={{\boldsymbol{Z}}_{m}^{k}}\odot {{\boldsymbol{D}}_{m}^{k}}
\end{aligned}
\end{equation}
\par After determining the specific retouching operation and corresponding parameter, we retouch the image to attack DNNs. However, the binary decision mask ${\boldsymbol{D}}_{m}^{k}$, obtained through sampling in Eq. \ref{eq6}, is not differentiable, which hinders end-to-end attack. To tackle this challenge, we employ the Gumbel-Softmax~\cite{jang2016categorical} technique to sample from the probability $\boldsymbol{P}_{m}^{k}$:
\begin{equation} \label{eq8}
\begin{aligned}
  {\boldsymbol{D}_{m}^{k}}=\text{Gumbel-Softmax}\left( {\boldsymbol{P}_{m}^{k}} \right)
\end{aligned}
\end{equation}
\par Gumbel-Softmax employs reparameterization trick to achieve differentiable sampling of discrete variables. In Eq.\ref{eq8}, the output from Gumbel-Softmax is a one-hot tensor of length $N$, equivalent to directly sampling from $\boldsymbol{P}_{m}^{k}$. With the differentiability of Gumbel-Softmax, we can update the action probability ${\boldsymbol{P}_{m}^{k}}$ and the retouching parameter ${{Z}_{m,n}^{k}}$ in table through backpropagation. 

\subsection{Optimization for Adversarial Retouching}\label{sec3.3}
\subsubsection{Retouching Style Guidance Module}\label{sec3.3.1}
\par
\begin{figure}
  \includegraphics[width=1\linewidth]{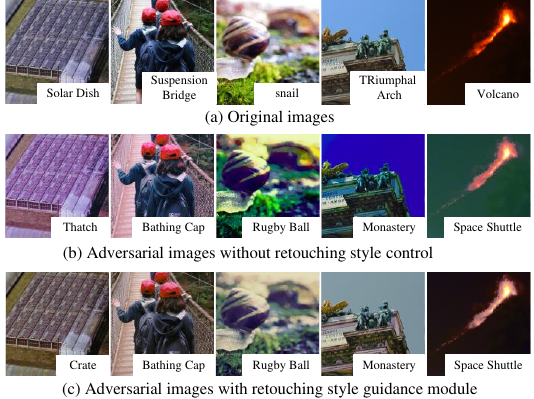}
  \caption{Visualization of original and adversarial images~(w/o and with retouching style control) generated by RetouchUAA on Inception V3. Classification labels at each image's bottom right.}
  \label{fig4}
  \vspace{-0.6cm}
\end{figure}

 The above attack framework ensures process-correct image retouching but lacks style control, leading to potential unrealistic and perceivable adversarial images~(as shown in Figure~\ref{fig4}b). This obstructs our analysis of the effects of retouching styles from real scenarios on DNNs, while also diminishing the stealth of the retouching style in attacks. To track this issue, we developed a retouching guidance module for more lifelike adversarial images. We opted not to use popular telegraph filters like AdvCF does due to their fixed style class and limited number of styles, which could restrict our exploration of retouching style risks. Instead, we utilized the MIT-Adobe FiveK Dataset~\cite{bychkovsky2011learning} with 25,000 professionally retouched images that are considered real-world standards, denoted as $\boldsymbol{x}^{std}$. 
To guide the adversarial image retouching style with that of these standards, we apply randomized retouching parameters, $\boldsymbol{\theta}^{*}$, modifying the style of the standard images to $\mathcal{R}(\boldsymbol{x}^{std}, \boldsymbol{\theta}^{*})$. Then, We then deploy a U-Net network with skip connections, $\mathcal{U}$, to compute the $L_2$ distance from $\mathcal{R}(\boldsymbol{x}^{std}, \boldsymbol{\theta}^{*})$ to the difference of standard image, which constitutes the style loss $\mathcal{L}_{style}$ for RetouchUAA.

\begin{equation} \label{eq9}
\begin{aligned}
{\mathcal{L}_{style}} = {{\left\| \mathcal{R}\left( {\boldsymbol{x}^{std}},{\boldsymbol{\theta }^{*}} \right)-{\boldsymbol{x}^{std}} \right\|}_{2}}=\mathcal{U}\left( \mathcal{R}\left( {\boldsymbol{x}^{std}},{\boldsymbol{\theta }^{*}} \right) \right)
\end{aligned}
\end{equation}
\par In optimization, we minimize the retouching style loss $\mathcal{L}_{style}$ to guide our attack framework produce standard-style adversarial images. Figure~\ref{fig4}c displays these images, more natural and realistic than those in Figure~\ref{fig4}b. Details of the $\mathcal{U}$ can be found in the supplementary material.

\subsubsection{Retouching Decision Regularization} \label{sec3.3.2}
In our attack framework, the Decision-Based Retouching Module (DRM) samples retouching operations from a probability table. We observed that DRM often favors certain operations, ignoring diverse combinations. This biased decision-making limits RetouchUAA's attack direction, compromising its attack ability. To counter this, we introduced a specific regularization $\mathcal{L}_{DRM}$ for DRM.
\begin{equation} \label{eq10}
\begin{aligned}
\mathcal{L}_{DRM} &= \mathcal{D}(\boldsymbol{S}_p || \boldsymbol{S}_q) = \sum_{o} \boldsymbol{S}_p(o) \log \frac{\boldsymbol{S}_p(o)}{\boldsymbol{S}_q(o)}
\end{aligned}
\end{equation}
where $\boldsymbol{S}_{p}$ denotes the distribution of retouching operation types sampled by DRM in an attack, and $\boldsymbol{S}_{q}$ is a discrete uniform distribution of the same dimension as $\boldsymbol{S}_{p}$. $\mathcal{D}$ denotes Kullback-Leibler divergence, and $o$ symbolizes a potential retouching operation type. To diversify retouching combinations, DRM is prompted to sample different operations by minimizing $\mathcal{L}_{DRM}$ during the attack.

\subsubsection{Persistent Attack Strategy} \label{sec3.3.3}
\par To control the perturbation magnitude and ensure imperceptibility, the most constrained iterative attacks usually stops and outputs an adversarial sample after the first successful attack. However, for unconstrained adversarial attacks, perturbation magnitude and imperceptibility are not explicitly correlated. Therefore, we design the persistent attack strategy for RetouchUAA. Following the initial attack's success, it continues with $I$ (we set $I=30$ in our experiment) iterations, driving the adversarial image further away from the decision boundary to improve its transferability and defense robustness. To balance imperceptibility, we choose the adversarial image with the smallest style loss ${\mathcal{L}_{style}}$ from the $I$ iterations as RetouchUAA’s output.

\subsection{Attacking Scheme} \label{sec3.4}
Jointly with the previously proposed attack framework and optimization strategy, the objective function of RetouchUAA in Eq. \ref{eq2} can be rewritten as
\begin{equation} \label{eq11}
\begin{aligned}
\underset{\boldsymbol{\theta} =\left[ \boldsymbol{P},\boldsymbol{Z} \right]}{\mathop{\arg \max }}\,\mathcal{J}\left( \mathcal{F}\left( \mathcal{C}\left( {\boldsymbol{x}^{real}},\boldsymbol{\theta} \right) \right),y \right)- {{\lambda }_{1}}{{\mathcal{L}}_{style}}-{\lambda_{DRM}{\lambda}_{2}}{{\mathcal{L}}_{DRM}}
\end{aligned}
\end{equation}
where $\boldsymbol{P}$ and $\boldsymbol{Z}$ denote the action probability table and retouch parameters, respectively. ${\lambda }{1}$ and ${\lambda }{2}$ are dynamic weights for regularization loss $\mathcal{L}{style}$ and $\mathcal{L}_{DRM}$, aligning with the visual task loss $\mathcal{J}\left( \mathcal{F}\left( \mathcal{C}\left( {\boldsymbol{x}^{real}}, \boldsymbol{\theta} \right) \right),y \right)$ in terms of magnitude. $\lambda_{DRM}$ further adjusts $\mathcal{L}_{DRM}$'s learning rate, set at 50 in our experiment.

\subsubsection{Implementation Details} \label{sec3.4.1}
We use PGD with an Adam optimizer featuring an incremental learning rate to solve Eq. \ref{eq11}. Initially, the learning rates for the action probability table $\boldsymbol{P}$ and corresponding retouching parameters $\boldsymbol{Z}$ are 1 and 0.0005, respectively. These learning rates increase linearly tenfold up to $Iter^{max}=1000$ iterations, aiding RetouchUAA generates finer retouched images without sacrificing attack strength. Additionally, we assign soft mask count $K$ and maximum action count $M$ as 5 and 30. Supplementary material details the impact of these hyperparameter adjustments on RetouchUAA's performance.

\begin{table*}[t]
    \centering
    \includegraphics[width=0.9\textwidth]{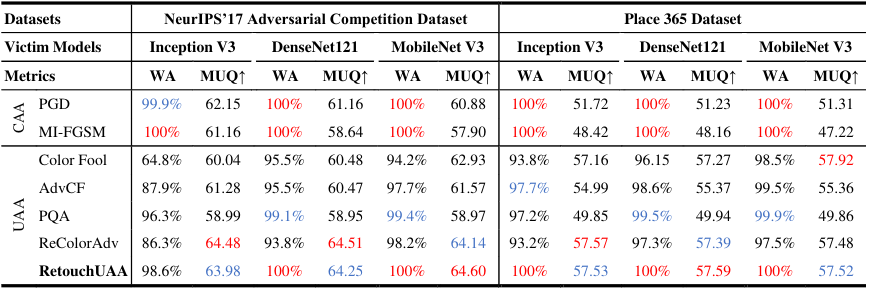}
    \caption{White-box attack success rate~(WA), and image naturalness~(MUQ) of RetouchUAA and baseline in the NeurIPS'17 adversarial competition dataset and Place 365 dataset. The arrow direction on each image quality metric indicates better naturalness. The optimal and suboptimal metrics are marked in \textcolor{myred}{red} and \textcolor{myblue}{blue}, respectively.}
    \label{tab1}
    \vspace{-0.4cm}
\end{table*}

\section{Experimental Results} \label{sec4}
In this section, we conduct a systematic analysis of the robustness of DNNs against retouching style from the adversarial attack perspective, covering imperceptibility, white-box and black-box attack strength and defense robustness. In addition, we compare RetouchUAA with existing attacks, especially unconstrained attacks, to explore the efficacy of retouching style as a potent attack tool.

\noindent \textbf{Datasets, Models, and Baselines.}
We investigated DNNs' robustness against RetouchUAA in object and scene classification tasks using the NeurIPS'17 adversarial competition dataset~\cite{kurakinnips2017} and the Place365 dataset~\cite{zhou2017places} (selecting the top 3 images per category) respectively. We employed three well-known pre-trained models, Inception V3 (Inc3)~\cite{szegedy2016rethinking}, MobileNet V3 (MN3)~\cite{howard2019searching}, and DenseNet121 (DN121)~\cite{huang2017densely}, as victim models in the attacks.
We selected PGD~\cite{madry2018towards} and MI-FGSM~\cite{dong2018boosting} from CAA as well as PQA~\cite{gaoscale}, ReColorAdv~\cite{laidlaw2019functional}, AdvCF~\cite{zhao2020adversarial2}, and Color Fool~\cite{shamsabadi2020colorfool} from UAA as baselines. The parameters for CAA and UAA were configured based on the original paper and \cite{kim2020torchattacks}, respectively. All experiments were conducted using Nvidia GeForce RTX 4090 GPU.

\noindent \textbf{Metric.} 
We evaluate white-box strength with attack success rates and black-box strength via transfer success rates. As with many unconstrained attack studies~\cite{gao2021advhaze,guo2020watch,tian2021ava,zhao2020adversarial2,laidlaw2019functional,shamsabadi2020colorfool,bhattad2019unrestricted}, we employ the no-reference image quality metric~(MUQ), to evaluate image naturalness as a measure of imperceptibility. Finally, we evaluated RetouchUAA and baselines' defense robustness against three strategies: JPEG compression~\cite{xu2017feature}, random scaling~\cite{xie2017mitigating}, and a purification module-based defense: DiffPure~\cite{nie2022diffusion}.

\subsection{Robustness Analysis of DNNs Against Attacks} \label{sec4.1}
\subsubsection{White-Box Attack and Image Naturalness} \label{sec4.1.1}
\par In this section, we examine DNNs vulnerability  to retouching styles by white-box attacks. We also assess the accuracy of RetouchUAA's retouching style simulations and their non-suspicious as an attack vector in terms of image naturalness.
As Table~\ref{tab1} shows, RetouchUAA demonstrates remarkable effectiveness in white-box attack scenarios, achieving or approaching a 100\% success rate against three different victim models, a rate that rivals constrained attack methods and significantly exceeds existing unconstrained attacks.
This result conclusively demonstrates that DNN models are particularly susceptible to variations in retouching style. Moreover, when compared with other forms of attacks, such as rain~\cite{zhai2022adversarial} or localized color modifications~\cite{shamsabadi2020colorfool}, DNN models' outputs are more profoundly affected by retouching style.
\par In terms of image naturalness, RetouchUAA also achieved satisfactory results, consistently achieving top or near-top naturalness ratings across all adversarial images from the victim models, highlighting its realistic retouching style simulation. 
In application scenarios, The success of RetouchUAA communicates a important message that attackers can induce users to generate adversarial retouching images by tampering with the camera imaging pipeline or image processing software. Therefore, the retouching style, as a non-suspicious potential form of attack, requires vigilance by security officers.
We also noticed that ReColorAdv generated the most natural-looking images most of the time. However, its utilization of an implicit color transformation function to alter image colors, which sacrifices both interpretability and guidance in the real world.

\begin{figure*}[t]
    \centering
    \includegraphics[width=1\textwidth]{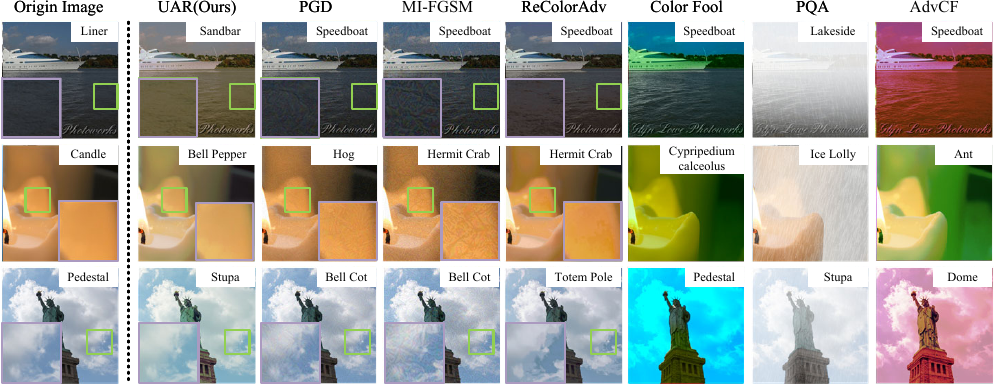}
    \caption{Adversarial examples generated by the baseline and RetouchUAA. The upper right corner of the image shows the label predicted by Inception V3, and the bottom-left corner shows a locally zoomed-in images, aiding in the comparison of certain perturbation details such as noise or posterization.}
    \label{fig5}
    \vspace{-0.0cm}
\end{figure*}

\begin{table*}[t]
    \centering
    \includegraphics[width=0.9\textwidth]{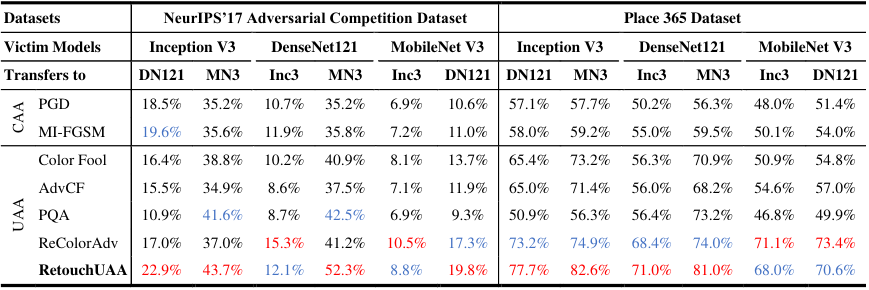}
    \caption{The transfer attack success rate of RetouchUAA and baseline in the NeurIPS'17 adversarial competition dataset and Place 365 dataset. 'Inc3', 'DN121', and 'MN3' denote Inception V3, DenseNet121, and MobileNet V3. The optimal and suboptimal metrics are marked in \textcolor{myred}{red} and \textcolor{myblue}{blue}, respectively.}
    \label{tab2}
    \vspace{-0.4cm}
\end{table*}

\par Beyond computing image quality evaluation metrics, Figure~\ref{fig5} also visualizes adversarial images from various attacks. 
Evidently, RetouchUAA generates more significant perturbations compared to PGD and MI-FGSM, owing to the $L_p$ norm constraints imposed on the latter. However, the images resulting from RetouchUAA maintain a more natural appearance, as its perturbations are relatively more subtle and harmonious, especially when contrasted with the noise introduced by PGD and MI-FGSM.
In addition, we also observed that ReColorAdv tends to introduce posterization, which compromises imperceptibility. Moreover, Color Fool generate unrealistic images due to their faulty segmentation or incorrect coloring. PAQ simulates climate change, but it may violate natural laws in some scenarios, like indoor scenes in the second row of Figure~\ref{fig5}. Finally, in the realm of unconstrained attacks, AdvCF is most similar to our attack, which simulates color control by adjusting color curves. However, in comparison to RetouchUAA, AdvCF struggles to precisely simulate adjustments to various image attributes(e.g., contrast and color temperature) and key retouching steps (e.g., image linearization, local retouching, and operation decisions). More critically, it adjusts image colors without adaptive guidance. These shortcomings limit its ability to effectively explore the impact of retouching styles on DNNs, along with its attack capabilities.

\begin{table*}[t]
    \centering
    \includegraphics[width=0.9\textwidth]{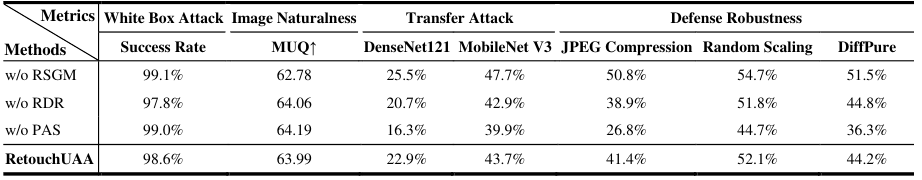}
    \caption{Ablation study of RetouchUAA on the NeurIPS'17 adversarial competition dataset with Inception V3 as the victim model.}
    \label{tab3}
    \vspace{-0.2cm}
\end{table*}

\subsubsection{Transferability and Defense Robustness} \label{sec4.1.2}
RetouchUAA's success in white-box attacks highlights DNNs' vulnerability to retouching style changes. More importantly, its retouching result is natural and non-suspicious, suggesting a potential attack vector. 
To further explore the threat of RetouchUAA, we conducted attack experiments in terms of black-box transferability as well as defense robustness. It should be noted that a trade-off exists between image naturalness and attack efficacy: pronounced perturbations enhance attack strength yet reduce naturalness. For a fair comparison, we adjusted the attack parameters to match RetouchUAA's naturalness~(MUQ), and then recorded their attack success rates after being transferred to other DNN models or defended by different defense strategies. Table~\ref{tab2} presents these transfer attack success rates.

\begin{figure}
 \vspace{-0.0cm}
  \includegraphics[width=1\linewidth]{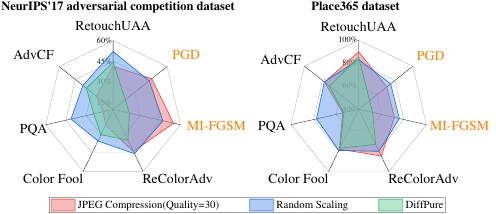}
  \caption{Defense robustness visual comparison with Inception V3 as victim model. Constrained attacks are marked in \textcolor{myorange}{orange} and the dataset used is highlighted in bold above each subplot.}
\vspace{-0.6cm}
  \label{fig6}
\end{figure}

\par As evidenced from Table~\ref{tab2}, it is observable that RetouchUAA generally surpasses the baseline in terms of transfer attack success rates, when the naturalness of the adversarial images are closely matched. Particularly noteworthy in the Place365 dataset, RetouchUAA maintains an attack success rate exceeding 65\% across various model transfers, even when transferring from weaker models like MobileNet V3 to stronger ones such as Inception V3. This underscores the robust transferability of RetouchUAA.
\par In terms of defense robustness, as shown in Figure~\ref{fig6}, RetouchUAA demonstrates satisfactory robustness against three prevalent defense strategies: JPEG compression, random scaling, and DiffPure. We observe that traditional defense tactics such as JPEG compression and random scaling have limited effectiveness in countering various attacks. DiffPure, a recently proposed purification defense, exhibits relatively superior performance, especially against attacks like PGD and MI-FGSM, attributed to its ability to effectively eliminate localized structural noise. However, RetouchUAA continues to exhibit robustness against DiffPure, securing impressive attack success rates even when subjected to this purification defense. We argue that this robustness of RetouchUAA is due to its tendency to generate spatially uniform noise, making it difficult to eliminate as noise.
In summary, our defense experiments reveal that many current defense strategies are ineffective due to the lack of specific design against retouch styles. Considering RetouchUAA's potent attack capabilities, tailored defenses against such attacks are necessary.

\subsection{Ablation Study}
\par In the following ablation experiment, we use Inception V3 as the victim model to examine the effectiveness of each block of RetouchUAA. Table~\ref{tab3} reports the metrics after respectively removing the following blocks: Retouching Style Guidance Module~(RSGM), Retouching Decision Regularization~(RDR) and Persistent Attack Strategy~(PAS).
\par First, the absence of RSGM in RetouchUAA leads to less image naturalness, as it doesn't restrict attack vectors, easily resulting in unnatural retouching.
We also note that RetouchUAA (w/o RSGM) shows improved attack performance, which is unsurprising given RSGM's style guidance limits on attack directions. Nonetheless, as RetouchUAA primarily aims to examine DNNs’ robustness to real-world retouching styles, preserving the naturalness of retouched images is vital.
Furthermore, it is observed that RDR enhances the attack strength and defense robustness of RetouchUAA by encouraging a combination of different retouching operations, while maintaining the naturalness of the image.
Finally, the inclusion of PAS slightly reduces image naturalness, likely due to its pursuit of more aggressive retouching styles. However, given its significant improvements to RetouchUAA's transferability and defense robustness, this slight naturalness trade-off is deemed acceptable.

\section{Conclusion} \label{sec5}
\par In this paper, we introduce a novel unconstrained adversarial retouching method, RetouchUAA, to study the impact of image retouching styles on DNNs from an adversarial attack perspective. 
By simulating human retouching behavior and retouching style guidance, RetouchUAA generates realistic adversarial retouched images and reveals the vulnerability of DNNs to the common real-world phenomenon: retouching styles.
Furthermore, with combined retouching decision regularization and persistent attack strategy, RetouchUAA not only achieves nearly 100\% white-box attack rate, but also strikes an optimal trade-off between image naturalness, transferability, and defense robustness compared to baselines, posing a significant threat to DNNs.
Our future work will be dedicated to exploring the feasibility of RetouchUAA as a black-box attack, as well as designing targeted defense strategies against such unconstrained attacks like RetouchUAA.

{
    \small
    \bibliographystyle{ieeenat_fullname}
    \bibliography{main}

\begin{thebibliography}{53}
\providecommand{\natexlab}[1]{#1}
\providecommand{\url}[1]{\texttt{#1}}
\expandafter\ifx\csname urlstyle\endcsname\relax
  \providecommand{\doi}[1]{doi: #1}\else
  \providecommand{\doi}{doi: \begingroup \urlstyle{rm}\Url}\fi

\bibitem[Afifi and Brown(2019)]{afifi2019else}
Mahmoud Afifi and Michael~S Brown.
\newblock What else can fool deep learning? addressing color constancy errors on deep neural network performance.
\newblock In \emph{IEEE International Conference on Computer Vision}, pages 243--252, 2019.

\bibitem[Afifi et~al.(2019)Afifi, Price, Cohen, and Brown]{afifi2019color}
Mahmoud Afifi, Brian Price, Scott Cohen, and Michael~S Brown.
\newblock When color constancy goes wrong: Correcting improperly white-balanced images.
\newblock In \emph{IEEE Conference on Computer Vision and Pattern Recognition}, pages 1535--1544, 2019.

\bibitem[Afifi et~al.(2022)Afifi, Abdelhamed, Abuolaim, Punnappurath, and Brown]{afifi2021cie}
Mahmoud Afifi, Abdelrahman Abdelhamed, Abdullah Abuolaim, Abhijith Punnappurath, and Michael~S Brown.
\newblock Cie xyz net: Unprocessing images for low-level computer vision tasks.
\newblock \emph{IEEE Transactions on Pattern Analysis and Machine Intelligence}, 44\penalty0 (9):\penalty0 4688–4700, 2022.

\bibitem[Athalye et~al.(2018)Athalye, Engstrom, Ilyas, and Kwok]{athalye2018synthesizing}
Anish Athalye, Logan Engstrom, Andrew Ilyas, and Kevin Kwok.
\newblock Synthesizing robust adversarial examples.
\newblock In \emph{International Conference on Machine Learning}, pages 284--293, 2018.

\bibitem[Bhattad et~al.(2020)Bhattad, Chong, Liang, Li, and Forsyth]{bhattad2019unrestricted}
Anand Bhattad, Min~Jin Chong, Kaizhao Liang, Bo Li, and David~A Forsyth.
\newblock Unrestricted adversarial examples via semantic manipulation.
\newblock In \emph{International Conference on Learning Representations}, 2020.

\bibitem[Bychkovsky et~al.(2011)Bychkovsky, Paris, Chan, and Durand]{bychkovsky2011learning}
Vladimir Bychkovsky, Sylvain Paris, Eric Chan, and Fr{\'e}do Durand.
\newblock Learning photographic global tonal adjustment with a database of input/output image pairs.
\newblock In \emph{IEEE Conference on Computer Vision and Pattern Recognition}, pages 97--104, 2011.

\bibitem[Carlini and Wagner(2017)]{carlini2017towards}
Nicholas Carlini and David Wagner.
\newblock Towards evaluating the robustness of neural networks.
\newblock In \emph{IEEE Symposium on Security and Privacy}, pages 39--57, 2017.

\bibitem[Chang et~al.(2015)Chang, Fried, Liu, DiVerdi, and Finkelstein]{chang2015palette}
Huiwen Chang, Ohad Fried, Yiming Liu, Stephen DiVerdi, and Adam Finkelstein.
\newblock Palette-based photo recoloring.
\newblock \emph{ACM Transactions on Graphics (Proc. SIGGRAPH)}, 34\penalty0 (4), 2015.

\bibitem[Dong et~al.(2018)Dong, Liao, Pang, Su, Zhu, Hu, and Li]{dong2018boosting}
Yinpeng Dong, Fangzhou Liao, Tianyu Pang, Hang Su, Jun Zhu, Xiaolin Hu, and Jianguo Li.
\newblock Boosting adversarial attacks with momentum.
\newblock In \emph{Proceedings of the IEEE conference on computer vision and pattern recognition}, pages 9185--9193, 2018.

\bibitem[Engstrom et~al.(2017)Engstrom, Tran, Tsipras, Schmidt, and Madry]{engstrom2017rotation}
Logan Engstrom, Brandon Tran, Dimitris Tsipras, Ludwig Schmidt, and Aleksander Madry.
\newblock A rotation and a translation suffice: Fooling cnns with simple transformations.
\newblock In \emph{arXiv preprint arXiv:1712.02779}, 2017.

\bibitem[Gao et~al.(2021)Gao, Guo, Juefei-Xu, Yu, and Feng]{gao2021advhaze}
Ruijun Gao, Qing Guo, Felix Juefei-Xu, Hongkai Yu, and Wei Feng.
\newblock Advhaze: Adversarial haze attack.
\newblock In \emph{arXiv preprint arXiv:2104.13673}, 2021.

\bibitem[Gao et~al.(2022)Gao, Luo, Lin, Xie, Liu, Shen, Kusumam, and Song]{gaoscale}
Xiangbo Gao, Cheng Luo, Qinliang Lin, Weicheng Xie, Minmin Liu, Linlin Shen, Keerthy Kusumam, and Siyang Song.
\newblock Scale-free and task-agnostic attack: Generating photo-realistic adversarial patterns with patch quilting generator.
\newblock In \emph{arXiv preprint arXiv:2208.06222}, 2022.

\bibitem[Goodfellow et~al.(2015)Goodfellow, Shlens, and Szegedy]{goodfellow2014explaining}
Ian~J Goodfellow, Jonathon Shlens, and Christian Szegedy.
\newblock Explaining and harnessing adversarial examples.
\newblock In \emph{International Conference on Learning Representations}, 2015.

\bibitem[Guo et~al.(2020)Guo, Juefei-Xu, Xie, Ma, Wang, Yu, Feng, and Liu]{guo2020watch}
Qing Guo, Felix Juefei-Xu, Xiaofei Xie, Lei Ma, Jian Wang, Bing Yu, Wei Feng, and Yang Liu.
\newblock Watch out! motion is blurring the vision of your deep neural networks.
\newblock In \emph{Advances in Neural Information Processing Systems}, pages 975--985, 2020.

\bibitem[He et~al.(2016)He, Zhang, Ren, and Sun]{he2016deep}
Kaiming He, Xiangyu Zhang, Shaoqing Ren, and Jian Sun.
\newblock Deep residual learning for image recognition.
\newblock In \emph{IEEE Conference on Computer Vision and Pattern Recognition}, pages 770--778, 2016.

\bibitem[Hosseini and Poovendran(2018)]{hosseini2018semantic}
Hossein Hosseini and Radha Poovendran.
\newblock Semantic adversarial examples.
\newblock In \emph{IEEE Conference on Computer Vision and Pattern Recognition Workshops}, pages 1614--1619, 2018.

\bibitem[Howard et~al.(2019)Howard, Sandler, Chu, Chen, Chen, Tan, Wang, Zhu, Pang, Vasudevan, et~al.]{howard2019searching}
Andrew Howard, Mark Sandler, Grace Chu, Liang-Chieh Chen, Bo Chen, Mingxing Tan, Weijun Wang, Yukun Zhu, Ruoming Pang, Vijay Vasudevan, et~al.
\newblock Searching for mobilenetv3.
\newblock In \emph{Proceedings of the IEEE/CVF international conference on computer vision}, pages 1314--1324, 2019.

\bibitem[Hu et~al.(2018)Hu, He, Xu, Wang, and Lin]{hu2018exposure}
Yuanming Hu, Hao He, Chenxi Xu, Baoyuan Wang, and Stephen Lin.
\newblock Exposure: A white-box photo post-processing framework.
\newblock \emph{ACM Transactions on Graphics (TOG)}, 37\penalty0 (4):\penalty0 1--15, 2018.

\bibitem[Huang et~al.(2017)Huang, Liu, Van Der~Maaten, and Weinberger]{huang2017densely}
Gao Huang, Zhuang Liu, Laurens Van Der~Maaten, and Kilian~Q Weinberger.
\newblock Densely connected convolutional networks.
\newblock In \emph{IEEE Conference on Computer Vision and Pattern Recognition}, pages 4700--4708, 2017.

\bibitem[Huang et~al.(2023)Huang, Sun, Guo, Juefei-Xu, Zhu, Feng, Liu, and Pu]{sun2022ala}
Yihao Huang, Liangru Sun, Qing Guo, Felix Juefei-Xu, Jiayi Zhu, Jincao Feng, Yang Liu, and Geguang Pu.
\newblock Ala: Naturalness-aware adversarial lightness attack.
\newblock In \emph{Proceedings of the 31st ACM International Conference on Multimedia}, pages 2418--2426, 2023.

\bibitem[Jang et~al.(2017)Jang, Gu, and Poole]{jang2016categorical}
Eric Jang, Shixiang Gu, and Ben Poole.
\newblock Categorical reparameterization with gumbel-softmax.
\newblock In \emph{International Conference on Learning Representations}, 2017.

\bibitem[Kanbak et~al.(2018)Kanbak, Moosavi-Dezfooli, and Frossard]{kanbak2018geometric}
Can Kanbak, Seyed-Mohsen Moosavi-Dezfooli, and Pascal Frossard.
\newblock Geometric robustness of deep networks: analysis and improvement.
\newblock In \emph{IEEE Conference on Computer Vision and Pattern Recognition}, pages 4441--4449, 2018.

\bibitem[Karaimer and Brown(2016)]{karaimer2016software}
Hakki~Can Karaimer and Michael~S Brown.
\newblock A software platform for manipulating the camera imaging pipeline.
\newblock In \emph{European Conference on Computer Vision}, pages 429--444, 2016.

\bibitem[Karras et~al.(2018)Karras, Aila, Laine, and Lehtinen]{karras2018progressive}
Tero Karras, Timo Aila, Samuli Laine, and Jaakko Lehtinen.
\newblock Progressive growing of gans for improved quality, stability, and variation.
\newblock In \emph{International Conference on Learning Representations}, 2018.

\bibitem[Kim(2020)]{kim2020torchattacks}
Hoki Kim.
\newblock Torchattacks: A pytorch repository for adversarial attacks.
\newblock In \emph{arXiv preprint arXiv:2010.01950}, 2020.

\bibitem[Kurakin et~al.(2017)Kurakin, Goodfellow, and Bengio]{kurakin2018adversarial}
Alexey Kurakin, Ian~J Goodfellow, and Samy Bengio.
\newblock Adversarial examples in the physical world.
\newblock In \emph{International Conference on Learning Representations}, pages 99--112, 2017.

\bibitem[Kurakin et~al.(2018)Kurakin, Goodfellow, Bengio, Dong, Liao, Liang, Pang, Zhu, Hu, Xie, et~al.]{kurakinnips2017}
Alexey Kurakin, Ian Goodfellow, Samy Bengio, Yinpeng Dong, Fangzhou Liao, Ming Liang, Tianyu Pang, Jun Zhu, Xiaolin Hu, Cihang Xie, et~al.
\newblock Adversarial attacks and defences competition.
\newblock In \emph{The NIPS'17 Competition: Building Intelligent Systems}, pages 195--231, 2018.

\bibitem[Laidlaw and Feizi(2019)]{laidlaw2019functional}
Cassidy Laidlaw and Soheil Feizi.
\newblock Functional adversarial attacks.
\newblock In \emph{Advances in neural information processing systems}, 2019.

\bibitem[Liu et~al.(2017)Liu, Ceylan, Yumer, Yang, and Lien]{liu2017material}
Guilin Liu, Duygu Ceylan, Ersin Yumer, Jimei Yang, and Jyh-Ming Lien.
\newblock Material editing using a physically based rendering network.
\newblock In \emph{IEEE International Conference on Computer Vision}, pages 2261--2269, 2017.

\bibitem[Liu et~al.(2022)Liu, Mao, Wu, Feichtenhofer, Darrell, and Xie]{liu2022convnet}
Zhuang Liu, Hanzi Mao, Chao-Yuan Wu, Christoph Feichtenhofer, Trevor Darrell, and Saining Xie.
\newblock A convnet for the 2020s.
\newblock In \emph{IEEE Conference on Computer Vision and Pattern Recognition}, pages 11976--11986, 2022.

\bibitem[Madry et~al.(2018)Madry, Makelov, Schmidt, Tsipras, and Vladu]{madry2018towards}
Aleksander Madry, Aleksandar Makelov, Ludwig Schmidt, Dimitris Tsipras, and Adrian Vladu.
\newblock Towards deep learning models resistant to adversarial attacks.
\newblock In \emph{International Conference on Learning Representations}, 2018.

\bibitem[Moran et~al.(2020)Moran, Marza, McDonagh, Parisot, and Slabaugh]{moran2020deeplpf}
Sean Moran, Pierre Marza, Steven McDonagh, Sarah Parisot, and Gregory Slabaugh.
\newblock Deeplpf: Deep local parametric filters for image enhancement.
\newblock In \emph{IEEE Conference on Computer Vision and Pattern Recognition}, pages 1052--1059, 2020.

\bibitem[Nguyen and Brown(2016)]{nguyen2016raw}
Rang~MH Nguyen and Michael~S Brown.
\newblock Raw image reconstruction using a self-contained srgb-jpeg image with only 64 kb overhead.
\newblock In \emph{IEEE Conference on Computer Vision and Pattern Recognition}, pages 1655--1663, 2016.

\bibitem[Nie et~al.(2022)Nie, Guo, Huang, Xiao, Vahdat, and Anandkumar]{nie2022diffusion}
Weili Nie, Brandon Guo, Yujia Huang, Chaowei Xiao, Arash Vahdat, and Anima Anandkumar.
\newblock Diffusion models for adversarial purification.
\newblock In \emph{International Conference on Machine Learning (ICML)}, 2022.

\bibitem[Qiu et~al.(2020)Qiu, Xiao, Yang, Yan, Lee, and Li]{qiu2020semanticadv}
Haonan Qiu, Chaowei Xiao, Lei Yang, Xinchen Yan, Honglak Lee, and Bo Li.
\newblock Semanticadv: Generating adversarial examples via attribute-conditioned image editing.
\newblock In \emph{European Conference on Computer Vision}, pages 19--37, 2020.

\bibitem[Rozsa et~al.(2016)Rozsa, Rudd, and Boult]{rozsa2016adversarial}
Andras Rozsa, Ethan~M Rudd, and Terrance~E Boult.
\newblock Adversarial diversity and hard positive generation.
\newblock In \emph{IEEE Conference on Computer Vision and Pattern Recognition Workshops}, pages 25--32, 2016.

\bibitem[Shamsabadi et~al.(2020)Shamsabadi, Sanchez-Matilla, and Cavallaro]{shamsabadi2020colorfool}
Ali~Shahin Shamsabadi, Ricardo Sanchez-Matilla, and Andrea Cavallaro.
\newblock Colorfool: Semantic adversarial colorization.
\newblock In \emph{IEEE Conference on Computer Vision and Pattern Recognition}, pages 1151--1160, 2020.

\bibitem[Song et~al.(2022)Song, Yu, Chen, and Yang]{song2022transformer}
Zikai Song, Junqing Yu, Yi-Ping~Phoebe Chen, and Wei Yang.
\newblock Transformer tracking with cyclic shifting window attention.
\newblock In \emph{IEEE conference on computer vision and pattern recognition}, pages 8791--8800, 2022.

\bibitem[Szegedy et~al.(2014)Szegedy, Zaremba, Sutskever, Bruna, Erhan, Goodfellow, and Fergus]{szegedy2013intriguing}
Christian Szegedy, Wojciech Zaremba, Ilya Sutskever, Joan Bruna, Dumitru Erhan, Ian Goodfellow, and Rob Fergus.
\newblock Intriguing properties of neural networks.
\newblock In \emph{International Conference on Learning Representations}, 2014.

\bibitem[Szegedy et~al.(2016)Szegedy, Vanhoucke, Ioffe, Shlens, and Wojna]{szegedy2016rethinking}
Christian Szegedy, Vincent Vanhoucke, Sergey Ioffe, Jon Shlens, and Zbigniew Wojna.
\newblock Rethinking the inception architecture for computer vision.
\newblock In \emph{IEEE Conference on Computer Vision and Pattern Recognition}, pages 2818--2826, 2016.

\bibitem[Tian et~al.(2021)Tian, Juefei-Xu, Guo, Xie, Li, and Liu]{tian2021ava}
Binyu Tian, Felix Juefei-Xu, Qing Guo, Xiaofei Xie, Xiaohong Li, and Yang Liu.
\newblock Ava: Adversarial vignetting attack against visual recognition.
\newblock In \emph{International Joint Conference on Artificial Intelligence}, 2021.

\bibitem[Wang et~al.(2004)Wang, Bovik, Sheikh, and Simoncelli]{wang2004image}
Zhou Wang, Alan~C Bovik, Hamid~R Sheikh, and Eero~P Simoncelli.
\newblock Image quality assessment: from error visibility to structural similarity regularized model.
\newblock \emph{IEEE transactions on image processing}, 13\penalty0 (4):\penalty0 600--612, 2004.

\bibitem[Xiao et~al.(2018)Xiao, Zhu, Li, He, Liu, and Song]{xiao2018spatially}
Chaowei Xiao, Jun-Yan Zhu, Bo Li, Warren He, Mingyan Liu, and Dawn Song.
\newblock Spatially transformed adversarial examples.
\newblock In \emph{International Conference on Learning Representations}, 2018.

\bibitem[Xie et~al.(2018)Xie, Wang, Zhang, Ren, and Yuille]{xie2017mitigating}
Cihang Xie, Jianyu Wang, Zhishuai Zhang, Zhou Ren, and Alan Yuille.
\newblock Mitigating adversarial effects through randomization.
\newblock In \emph{International Conference on Learning Representations}, 2018.

\bibitem[Xu et~al.(2018)Xu, Evans, and Qi]{xu2017feature}
Weilin Xu, David Evans, and Yanjun Qi.
\newblock Feature squeezing: Detecting adversarial examples in deep neural networks.
\newblock In \emph{25th Annual Network and Distributed System Security Symposium, NDSS 2018}, 2018.

\bibitem[Yan et~al.(2014)Yan, Lin, Kang, and Tang]{yan2014learning}
Jianzhou Yan, Stephen Lin, Sing~Bing Kang, and Xiaoou Tang.
\newblock A learning-to-rank approach for image color enhancement.
\newblock In \emph{IEEE Conference on Computer Vision and Pattern Recognition}, pages 2987--2994, 2014.

\bibitem[Zhai et~al.(2022)Zhai, Juefei-Xu, Guo, Xie, Ma, Feng, Qin, and Liu]{zhai2022adversarial}
Liming Zhai, Felix Juefei-Xu, Qing Guo, Xiaofei Xie, Lei Ma, Wei Feng, Shengchao Qin, and Yang Liu.
\newblock Adversarial rain attack and defensive deraining for dnn perception.
\newblock In \emph{arXiv preprint arXiv:2009.09205}, 2022.

\bibitem[Zhang et~al.(2021)Zhang, Guo, Gao, Juefei-Xu, Yu, and Feng]{zhang2021adversarial}
Qian Zhang, Qing Guo, Ruijun Gao, Felix Juefei-Xu, Hongkai Yu, and Wei Feng.
\newblock Adversarial relighting against face recognition.
\newblock In \emph{arXiv preprint arXiv:2108.07920}, 2021.

\bibitem[Zhao et~al.(2020{\natexlab{a}})Zhao, Liu, and Larson]{zhao2020adversarial1}
Zhengyu Zhao, Zhuoran Liu, and Martha Larson.
\newblock Adversarial color enhancement: Generating unrestricted adversarial images by optimizing a color filter.
\newblock In \emph{31st British Machine Vision Conference 2020, {BMVC} 2020}, 2020{\natexlab{a}}.

\bibitem[Zhao et~al.(2020{\natexlab{b}})Zhao, Liu, and Larson]{zhao2020towards}
Zhengyu Zhao, Zhuoran Liu, and Martha Larson.
\newblock Towards large yet imperceptible adversarial image perturbations with perceptual color distance.
\newblock In \emph{IEEE Conference on Computer Vision and Pattern Recognition}, pages 1039--1048, 2020{\natexlab{b}}.

\bibitem[Zhao et~al.(2023)Zhao, Liu, and Larson]{zhao2020adversarial2}
Zhengyu Zhao, Zhuoran Liu, and Martha Larson.
\newblock Adversarial image color transformations in explicit color filter space.
\newblock \emph{IEEE Transactions on Information Forensics and Security}, 18:\penalty0 3185--3197, 2023.

\bibitem[Zhong et~al.(2022)Zhong, Liu, Zhai, Jiang, and Ji]{zhong2022shadows}
Yiqi Zhong, Xianming Liu, Deming Zhai, Junjun Jiang, and Xiangyang Ji.
\newblock Shadows can be dangerous: Stealthy and effective physical-world adversarial attack by natural phenomenon.
\newblock In \emph{IEEE Conference on Computer Vision and Pattern Recognition}, pages 15345--15354, 2022.

\bibitem[Zhou et~al.(2017)Zhou, Lapedriza, Khosla, Oliva, and Torralba]{zhou2017places}
Bolei Zhou, Agata Lapedriza, Aditya Khosla, Aude Oliva, and Antonio Torralba.
\newblock Places: A 10 million image database for scene recognition.
\newblock \emph{IEEE Transactions on Pattern Analysis and Machine Intelligence}, 2017.

\end{thebibliography}
}


\end{document}


\vspace{10mm} 
\begin{center}
    \Large \textbf{Supplementary Material}
\end{center}
\vspace{0mm} 

\section{Differentiable Retouching Operation}\label{sec1}
\par We present the retouching-based attack framework in Section 3.2 of the manuscript, which generates retouched images for attacking DNNs by sampling differentiable retouching operations. The specific retouching operations we used are listed in Table~\ref{tabs1}.
\begin{table*}[h]
    \centering
    \includegraphics[width=0.7\textwidth]{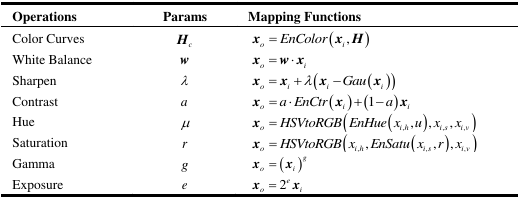}
    \caption{Parameter range and mapping function of differentiable retouching operations}
    \label{tabs1}
    \vspace{-0.2cm}
\end{table*}
\par \noindent where $\boldsymbol{x}_i$ and $\boldsymbol{x}_o$ represent the pixel values of the image before and after the retouch operation, respectively. $Gau$ represents Gaussian filtering, . ${x}_{i,h}$, ${x}_{i,s}$, and ${x}_{i,v}$ are the hue, saturation, and value channels of the input pixel $\boldsymbol{x}_i$ in the HSV color space. $HSVtoRGB$ is a function that performs the conversion from the HSV color space to the RGB color space. We describe the function details of Color Curve, Contrast, Hue and Saturation further below.

\subsection{Color Curve}\label{sec1.1}
In the color curve adjustment of the image, we followed the approach of Hu et al.~\cite{hu2018exposure} and designed the mapping function for the color curve as a monotonically piecewise linear function to make it differentiable. In detail, we set the parameters $\boldsymbol{H}$ for the piecewise linear function as ${\boldsymbol{H}}=( {{h}_{c,1}},{{h}_{c,2}},\ldots ,{{h}_{c,L-1}} )$ with $c\in \{ r,g,b \}$, and the points on this function are represented as $\left( {j}/{L},{{{T_{c,j}}}}/{{{T_{c,L}}}}\right)$, where ${T}_{c,j}=\sum\nolimits_{l=0}^{j-1}{{{h}_{c,l}}}$. Based on the above representation, the input pixel ${x}_{i,c}$ can be mapped to:
\begin{equation} \label{eqs1}
\begin{aligned}
  \boldsymbol{{x}_{o}} = EnColor(\boldsymbol{{x}_{i}},\boldsymbol{H}) = \frac{1}{T_{c,L}} \sum_{j=0}^{L-1} \text{clip}(L\cdot{{x}_{i,c}}-j,0,1) \cdot h_{c,j}
\end{aligned}
\end{equation}
We set $L$ in Eq.\ref{eqs1} to 64 following Zhao et al.~\cite{zhao2020adversarial2}

\subsection{Contrast}
\begin{equation} \label{eqs2}
\begin{aligned}
    \boldsymbol{{x}_{o}} = a \cdot EnCtr\left( {\boldsymbol{x}_{i}} \right)+\left( 1-a \right) \cdot {\boldsymbol{x}_{i}}
\end{aligned}
\end{equation}
\par for $EnCtr\left( {\boldsymbol{x}_{i}} \right)$:
\begin{equation} \label{eqs3}
\begin{aligned}
EnCtr\left( {\boldsymbol{x}_{i}} \right)={\boldsymbol{x}_{i}}\cdot \frac{EnLum\left( {\boldsymbol{x}_{i}} \right)}{Lum\left( {\boldsymbol{x}_{i}} \right)}
\end{aligned}
\end{equation}
\begin{equation} \label{eqs4}
\begin{aligned}
EnLum\left( {\boldsymbol{x}_{i}} \right)=\frac{1}{2}\left( 1-\cos \left( \pi \cdot Lum\left( {\boldsymbol{x}_{i}} \right) \right) \right)
\end{aligned}
\end{equation}
\par where $Lum\left( {\boldsymbol{x}_{i}} \right)=0.27{{x}_{i,r}}+0.67{{x}_{i,g}}+0.06{{x}_{i,b}}$

\subsection{Hue}
\begin{equation} \label{eqs5}
\begin{aligned}
\boldsymbol{{x}_{o}} = HSVtoRGB\left( EnHue\left( {{x}_{i,h}},u \right),{{x}_{i,s}},{{x}_{i,v}} \right)
\end{aligned}
\end{equation}
\par for $EnHue\left( {{x}_{i,h}},u \right)$:
\begin{equation} \label{eqs6}
\begin{aligned}
EnHue\left( {{x}_{i,h}},u \right)=\frac{{{x}_{i,h}}+u}{2\pi }
\end{aligned}
\end{equation}

\subsection{Saturation}
\begin{equation} \label{eqs7}
\begin{aligned}
\boldsymbol{{x}_{o}} = HSVtoRGB\left( {{x}_{i,h}},EnSatu\left( {{x}_{i,s}},r \right),{{x}_{i,v}} \right)
\end{aligned}
\end{equation}
\par for $EnSatu\left( {{x}_{i,s}},r \right)$:
\begin{equation} \label{eqs8}
\begin{aligned}
EnSatu\left( {{x}_{i,s}},r \right)=r\cdot {{x}_{i,s}}
\end{aligned}
\end{equation}

\section{Conversion from Linear sRGB to Nonlinear sRGB}\label{sec2}
\par In section 3.2.1 of the manuscript, we linearized the image by applying inverse gamma correction. Here, we restore the image to a nonlinear state by reapplying gamma correction for the purpose of visualizing and deceiving the classification model.
\begin{equation} \label{eqs9}
\begin{aligned}
\boldsymbol{x}^{NL}=\left\{ \begin{matrix}
   {\boldsymbol{x}^{L}} \times {12.92}\; & if\ \boldsymbol{x}^{NL}\le 0.00313  \\
   1.055\times {{\left( \boldsymbol{x}^{L} \right)}^{\frac{1}{2.4}}}-0.055 & otherwise  \\
\end{matrix} \right\} \\
\end{aligned}
\end{equation}

\section{Hyper Parameter Analysis}\label{sec3}
\par This section investigates the effects of different hyper parameter settings on the performance of RetouchUAA. We consider the following key hyper parameters in RetouchUAA:
\begin{itemize}
    \item 
    The number of palette-based masks $K$ generated in local retouching and the number of actions $M$ in the Decision-Based Retouching module (DRM).
    \item 
    The learning rate ${\ell}_{Z}$ for the retouch parameters and ${\ell}_{P}$ for the sampling probability of the retouch action.
    \item 
    The number of iterations $I$ in persistent attack strategy.
    \item 
    The $\lambda_{DRM}$ from Eq. 11 in manuscript, which is used to adjust the learning rate for the regularization loss ${\mathcal{L}}_{DRM}$
\end{itemize}

\subsection{The Number of Palette-Based Masks $K$ and Actions $M$}\label{sec3.1}
\vspace{-0.2cm}
\begin{figure}[h]
  \centering
  \includegraphics[width=0.9\textwidth]{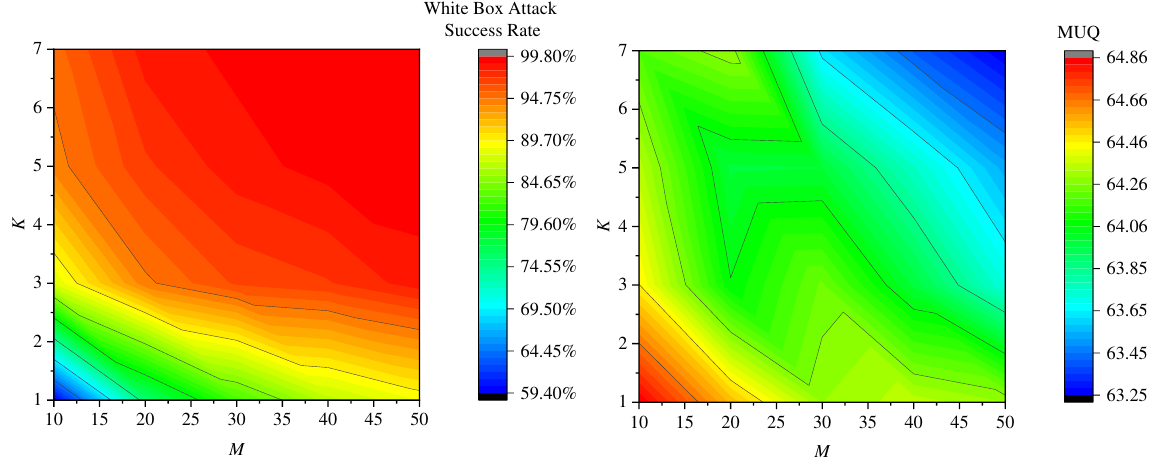}
  \vspace{-0.2cm}
  \caption{Impact of mask~($K$) and action~($M$) numbers on RetouchUAA performance, including white-box attack success rate and image naturalness~(MUQ).}
  \label{figs1}
  \vspace{-0.4cm}
\end{figure}
\par As depicted in Figure~\ref{figs1}, the success rate of RetouchUAA's white-box attacks progressively increases with the augmentation of the $K$ and $M$. Simultaneously, image naturalness~(MUQ) correspondingly declines. This is consistent with our expectations, as increasing $K$ and $M$ provides a board range of choices for retouch-based attacks, but due to the unconstrained nature of RetouchUAA, it could also lead to excessive retouching, thereby diminishing the naturalness of the images. Additionally, we observe an interesting fact that, compared to the number of actions $M$, increasing the number of masks $K$ significantly improves the attack capability of RetouchUAA without sacrificing too much naturalness. For instance, when $M=30$, increasing $K$ boosts the success rate of the white-box attack from 75\% to nearly 100\%, but there is no significant change in the naturalness of the retouched images. Based on this observation, we set $M=30$ and $K=5$ in our experiments.

\subsection{The Learning Rate of the Retouching Parameter ${\ell}_{Z}$ and of the Action Sampling Probability ${\ell}_{P}$}\label{sec3.2}
\begin{figure}[h]
  \centering
  \includegraphics[width=0.9\textwidth]{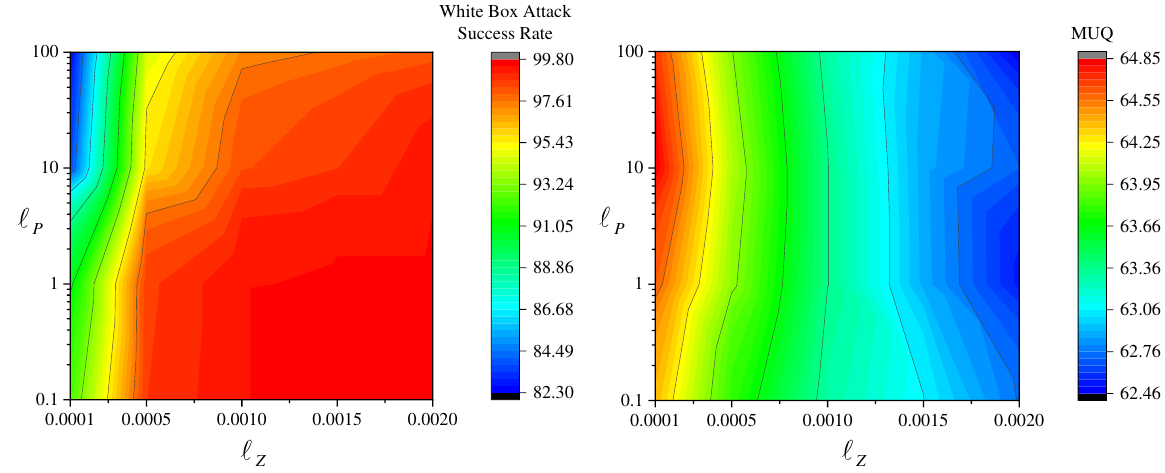}
  \vspace{-0.2cm}
  \captionof{figure}{Impact of the learning rate of the retouching parameter ${\ell}_{Z}$ and of the action sampling probability ${\ell}_{P}$ on RetouchUAA performance, including white-box attack success rate and image naturalness~(MUQ).}
  \label{figs2}
  \vspace{-0.5cm}
\end{figure}
\par Figure~\ref{figs2} shows that RetouchUAA has a higher success rate in white-box attacks when $\ell_{P}$ is set to smaller values, such as 1 or 0.1, suggesting that a lower $\ell_{P}$ is beneficial for RetouchUAA to identify more threatening combinations of retouching operations. Differently, the impact of variations in $\ell_{P}$ on images naturalness~(MUQ) appears to be negligible.
Contrary to $\ell_{P}$, an increase in $\ell_{Z}$ leads to a marked improvement in the success rate of RetouchUAA's white-box attacks, though this comes at the expense of reduced images naturalness. This pattern can be attributed to the fact that higher ${\ell}_{Z}$ correspond to more significant retouching style alterations, which can consequently generate more aggressive and possibly less natural retouching styles. Based on these observations, we opted for ${\ell}_{P}=1$ and ${\ell}_{Z}=0.0005$ in our experiments.

\subsection{Persistent Attack Iteration Number $I$}\label{sec3.3}
\vspace{0.2cm}
\begin{center}
    \includegraphics[width=0.9\textwidth]{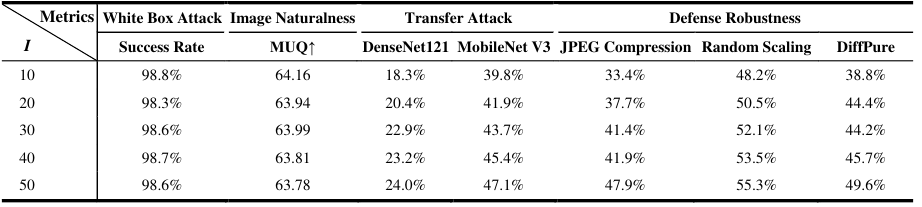}
    \vspace{-0.2cm}
    \captionof{table}{Impact of persistent attack iteration number $I$ on RetouchUAA performance.}
    \label{tabs2}   
\end{center}
 \vspace{-0.1cm}
\par As shown in Table~\ref{tabs2}, following the initial successful attack, the transferability and defense robustness of RetouchUAA significantly improve with an increasing number of iterations in the persistent attack, at the cost of only a slight decrease in the naturalness of the images. This is attributed to RetouchUAA maintaining the naturalness of the image during the persistent attack by selecting retouched images with the minimal style loss, while pushing these images further away from the decision boundary of the original label. Considering the trade-off between the naturalness and attack strength of RetouchUAA, we set ${I}=30$ in our experiments.

\subsection{The Learning Rate $\lambda_{DRM}$ of DRM's Regularization Loss}\label{sec3.4}
\vspace{0.2cm}
\begin{center}
    \includegraphics[width=0.9\textwidth]{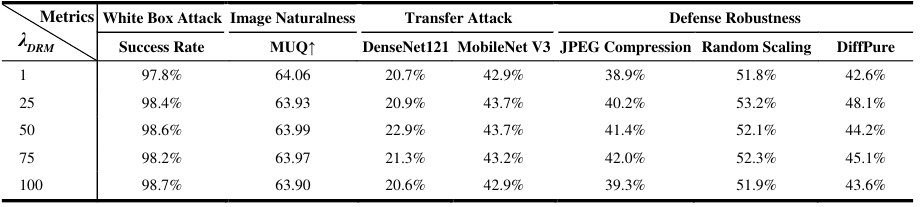}
    \vspace{-0.2cm}
    \captionof{table}{Impact of learning rate $\lambda_{DRM}$ on RetouchUAA performance.}
    \label{tabs3}
\end{center}
\par  As can be seen in Table~\ref{tabs3}, most of RetouchUAA's metrics, especially transferability and defensive robustness, show an initial improvement followed by a decline with escalation of $\lambda_{DRM}$. This result suggests that an appropriate $\lambda_{DRM}$ helps to suppress the tendency of the DRM to select certain fixed retouching operations, thus obtaining a more diversified combination of retouching operations to improve the attack capability. However, an excessively large $\lambda_{DRM}$ will cause the regularized loss ${\mathcal{L}}_{DRM}$ to become dominant, leading RetouchUAA to overlook the optimization of other loss terms, such as visual task loss, thereby impairing the performance of RetouchUAA. Based on this observation, we set $\lambda_{DRM}=50$ in our experiments.

\section{Network Architecture for Style Guide Modules}\label{sec4}
\par We utilize a U-Net network architecture with skip connections to generate pixel-level style losses for adversarial retouched images, as shown in Table~\ref{tabs5}. We train this network for 500 epochs using an Adam optimizer with a learning rate of 0.001.
\begin{table*}[h]
    \centering
    \includegraphics[width=1\textwidth]{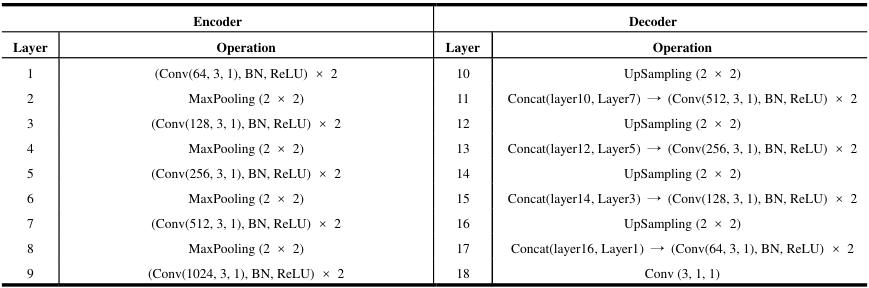}
    \captionof{table}{The network architecture for style guidance}
    \label{tabs5}
    \vspace{-0.2cm}
\end{table*}
\par \noindent where 'Conv' stands for convolution operation, and 'Conv(64,3,1)' means a convolution layer with a stride of 1 using a $3\times3$ convolution to output 64 feature maps. 'BN' stands for Batch Normalization, and 'ReLU' represents the activation function. 'MaxPooling' and 'UpSampling' respectively denote the downsampling and upsampling operations, with a scaling factor of $2\times2$. 'Concat' is used to concatenate feature maps from two different layers. Further network architecture optimization may contribute to style loss accuracy.

\section{Visual Explanation of RetouchUAA}
\par
\begin{figure}[H]
  \centering
  \includegraphics[width=0.85\textwidth]{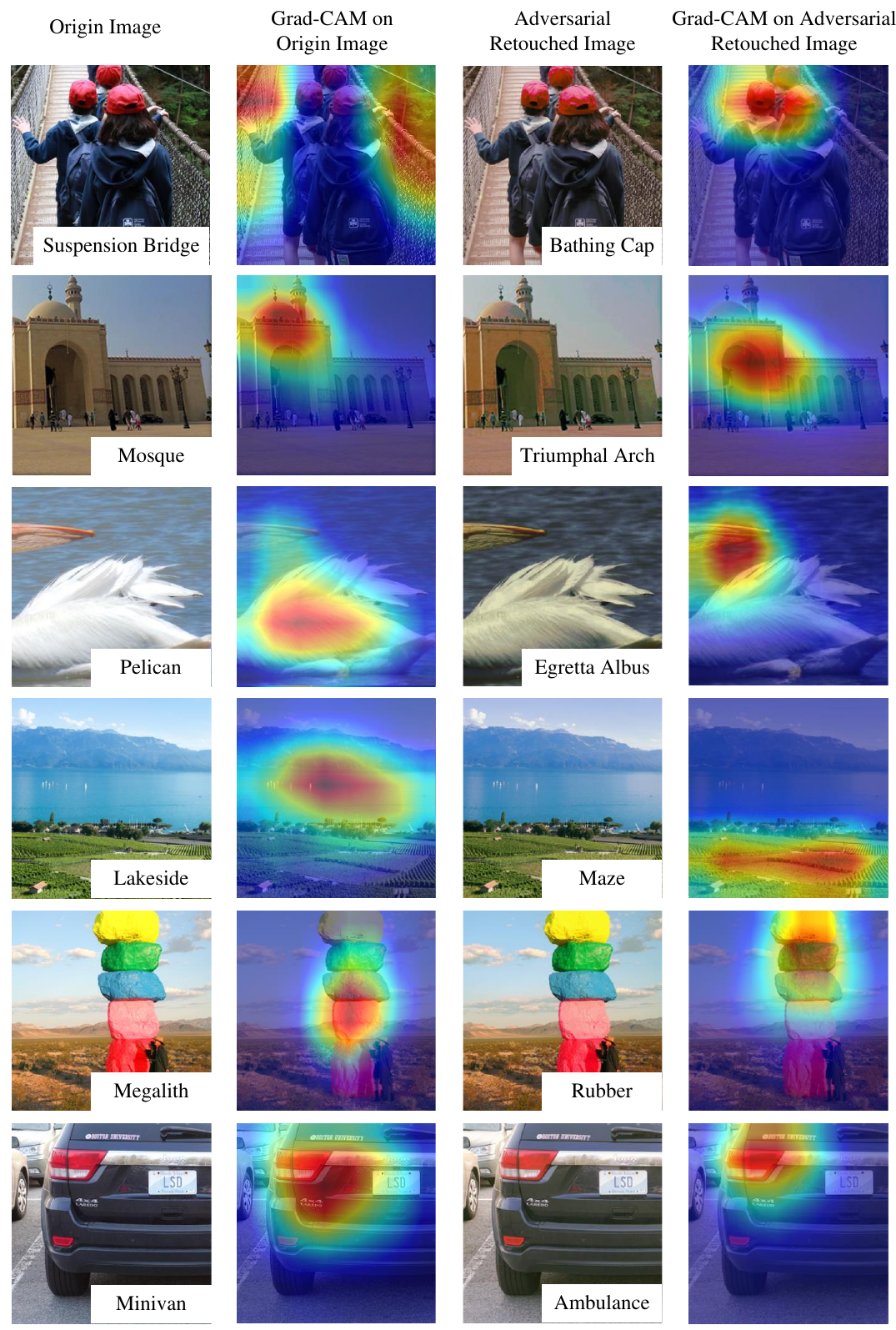}
  \caption{Heatmaps of the original image and the adversarial retouched image generated by Grad-CAM, based on the Inception-v3~(Mixed-7c layer). The red areas in the heatmap indicate the model's focus. Ground truth and model predicted labels are respectively marked in the bottom right corner of the original and retouched images.}
  \label{figs3}
  \vspace{-0.4cm}
\end{figure}

 To visually understand how RetouchUAA deceives the model, we employed Grad-CAM~\cite{selvaraju2017grad} to visualize the heatmaps of the Mixed-7c layer in Inception-v3 for both the original image and the adversarial retouched image generated by RetouchUAA in Figure~\ref{figs3}.
It is evident that the first three rows of images in Figure~\ref{figs3} exhibit significant yet natural perturbations, which noticeably alter the model's focus regions and result in incorrect classification. For instance, in the first row, the model’s focus shifts from the bridge to the human body due to the change in retouching style, resulting in a discrepancy from the prediction of the original image. Likewise, the variations in retouching style among the images in the second and third rows of Figure~\ref{figs3} also modify the model's focus regions. Although this does not cause the model's focus to deviate from the target, it successfully deceives the model by reducing the emphasis on crucial components of the target, such as the dome structure in the mosque and the beak in the Pelican.
Additionally, we have observed that even minor changes in retouching style can significantly disrupt the model's focus regions. Figure~\ref{figs3} rows 4 to 6 show that the retouching style changes for these adversarial images are barely noticeable compared to the original images. However, the model’s focus regions differ considerably, indicating the sensitivity of DNNs to retouching style variations.
\newpage
\newpage

\section{Visualization of Adversarial Retouched Images}
\begin{figure}[H]
  \centering
  \includegraphics[width=0.74\textwidth]{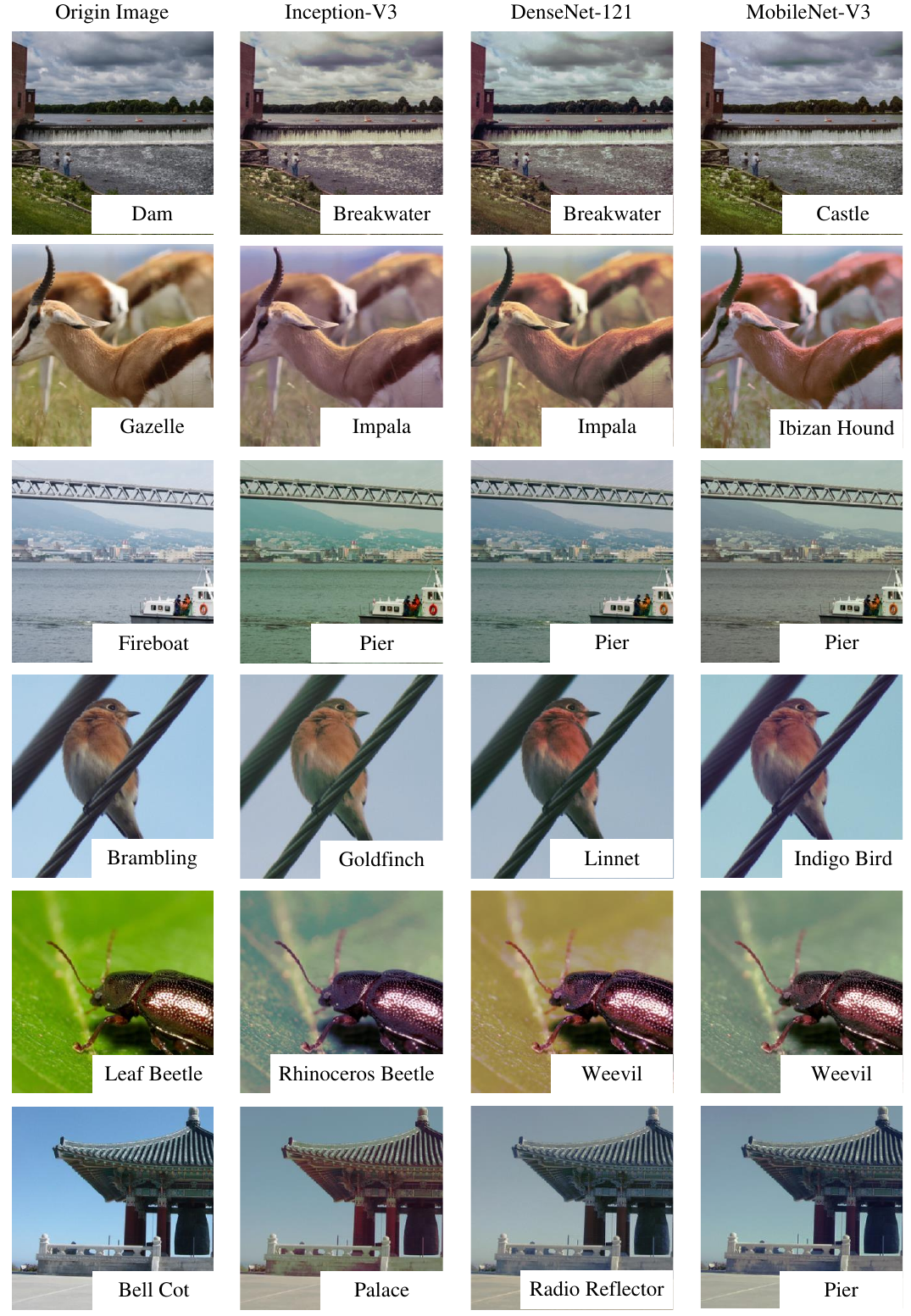}
  \caption{Additional adversarial retouched images generated by RetouchUAA for Inception-V3, DenseNet-121 and MobileNet-V3 in the NeurIPS'17 adversarial competition dataset. Ground truth and predicted labels from different models are respectively marked in the bottom right corner of the original image and the corresponding adversarial retouched image.}
  \label{figs4}
  \vspace{-0.4cm}
\end{figure}

\begin{figure}[H]
  \centering
  \includegraphics[width=0.85\textwidth]{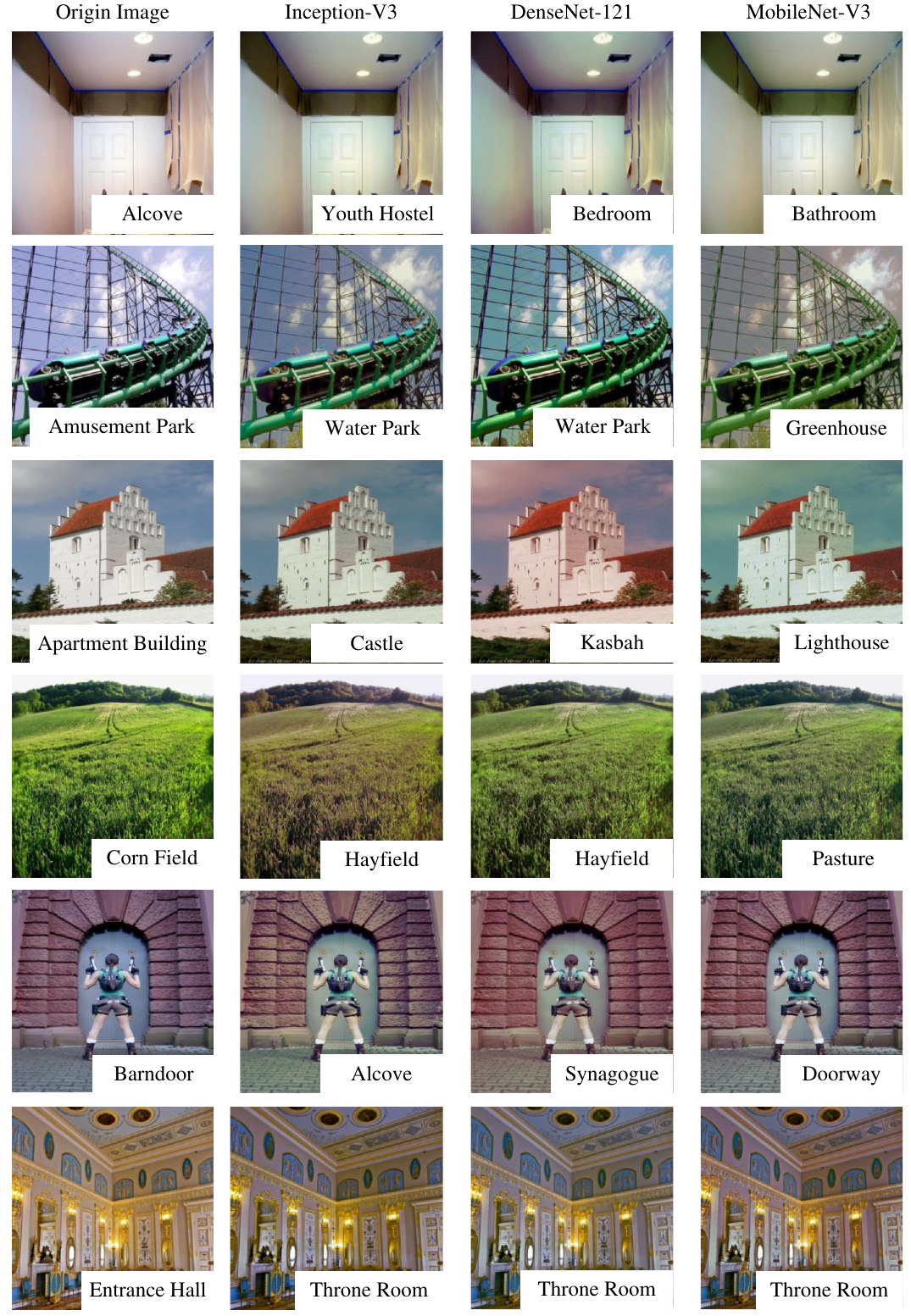}
  \caption{Additional adversarial retouched images generated by RetouchUAA for Inception-V3, DenseNet-121 and MobileNet-V3 classification models in the Place365 dataset. Groundtruth annotations and predicted labels from different models are respectively marked in the bottom right corner of the original image and the corresponding adversarial retouched image.}
  \label{figs5}
  \vspace{-0.4cm}
\end{figure}

We provide additional retouched images from the NeurIPS'17 adversarial competition dataset~\cite{kurakinnips2017} and the Place365 dataset~\cite{ zhou2017places} in Figure~\ref{figs4} and Figure~\ref{figs5} to evalute the naturalness of the adversarial retouched images generated by RetouchUAA. These images are created by RetouchUAA to attack Inception-V3, DenseNet-121 and MobileNet-V3 in the white-box attack scenario, respectively.

\newpage

\newpage

{
    \small
    \bibliographystyle{ieeenat_fullname}
    \bibliography{main}
}
